\documentclass[10pt,twocolumn,letterpaper]{article}

\usepackage{cvpr}              % To produce the CAMERA-READY version
\definecolor{cvprblue}{rgb}{0.21,0.49,0.74}
\usepackage[pagebackref,breaklinks,colorlinks,allcolors=cvprblue]{hyperref}
\usepackage[dvipsnames]{xcolor}
\usepackage{array}
\usepackage[T1]{fontenc}

\usepackage{booktabs}
\usepackage{threeparttable}
\usepackage{titletoc}   % do NOT touch hyperref

\makeatletter
\newcommand{\myfnsymbol}[1]{%
  \ifcase#1 *\or \dagger\or \ddagger\or \S\or \P\or \|\or **\or \dagger\dagger\or \ddagger\ddagger\else *\fi}
\makeatother

\newcommand{\tabfn}[1][]{%
  \textsuperscript{%
    \if\relax\detokenize{#1}\relax *\else \myfnsymbol{#1}\fi}}

\def\paperID{17359} % *** Enter the Paper ID here
\def\confName{CVPR}
\def\confYear{2026}
\def\dd{\mathrm{d}}
\def\E{\mathbb{E}}
\def\Ls{\mathcal{L}}
\def\sR{\mathbb{R}}
\def\gN{\mathcal{N}}

\def\gY{\mathcal{Y}}
\def\sbfm/{Schr\"{o}dinger Bridge Flow Matching}

\title{Schr\"{o}dinger Audio-Visual Editor: Object-Level Audiovisual Removal}

\author{
\parbox{\linewidth}{\centering
Weihan Xu$^{1}$\thanks{Equal contribution.}\quad
Kan Jen Cheng$^{2,3}$\footnotemark[1]\quad
Koichi Saito$^{4}$\quad
Muhammad Jehanzeb Mirza$^{1}$\\
Tingle Li$^{2,3}$\quad
Yisi Liu$^{2,3}$\quad
Alexander Liu$^{1}$\quad
Liming Wang$^{1}$\quad
Masato Ishii$^{4}$\\
Takashi Shibuya$^{4}$\quad
Yuki Mitsufuji$^{4,5}$\quad
Gopala Anumanchipalli$^{2,3}$\quad
Paul Pu Liang$^{1}$\\[0.4em]
$^{1}$MIT\quad
$^{2}$UC Berkeley\quad
$^{3}$Berkeley AI Research\quad
$^{4}$Sony AI\quad
$^{5}$Sony Group Corporation
}
}

\begin{document}
\maketitle
\newcommand{\tba}[1][]{\textcolor{blue}{TBA\ifx&#1&\else---#1\fi}}
\definecolor{darkgreen}{rgb}{0.0,0.5,0.0}
\definecolor{darkred}{rgb}{0.6,0.0,0.0}
\newcommand{\yes}{\textcolor{darkgreen}{\checkmark}}
\newcommand{\no}{\textcolor{darkred}{$\times$}}
\begin{abstract}

Joint editing of audio and visual content is crucial for precise and controllable content creation. This new task poses challenges due to the limitations of paired audio-visual data before and after targeted edits, and the heterogeneity across modalities. To address the data and modeling challenges in joint audio-visual editing, we introduce SAVEBench, a paired audiovisual dataset with text and mask conditions to enable object-grounded source-to-target learning. With SAVEBench, we train the Schr\"{o}dinger Audio-Visual Editor (SAVE), an end-to-end flow-matching model that edits audio and video in parallel while keeping them aligned throughout processing. SAVE incorporates a Schr\"{o}dinger Bridge that learns a direct transport from source to target audiovisual mixtures. Our evaluation demonstrates that the proposed SAVE model is able to remove the target objects in audio and visual content while preserving the remaining content, with stronger temporal synchronization and audiovisual semantic correspondence compared with pairwise combinations of an audio editor and a video editor.
\end{abstract}    
\section{Introduction}
\label{sec:intro}

Modern content creation demands precise, controllable editing across modalities. For instance, in an Antarctic wildlife shoot, an unexpected seal call may dominate the audio just as it enters a scene, requiring coordinated audio–visual editing to prevent a costly reshoot. Enabling this kind of correction requires an end-to-end system that edits audio and video jointly given natural-language instructions while maintaining close cross-modal alignment. In contrast to simply combining independent audio~\cite{wang2023audit, manor2024zeroshotunsupervisedtextbasedaudio} and video editors~\cite{zhou2023propainterimprovingpropagationtransformer, jiang2025vaceallinonevideocreation, li2022endtoendframeworkflowguidedvideo}, joint editing can better preserve temporal audiovisual synchronization and semantic alignment because the modalities naturally learn from each other. In addition, natural language-guided audio-visual editing is more expressive, precise, and intuitive, enabling edits without extra exemplars or scene descriptions~\cite{brooks2023instructpix2pixlearningfollowimage,correia2021avui, Perazzi2016DAVIS}.

However, joint audio-visual editing is challenging since it requires training on paired corpora of videos with aligned audio~\cite{audiovisual_survey, lee2021acav100mautomaticcurationlargescale}. This makes audiovisual editors substantially more data-hungry in practice than unimodal ones. Curating such corpora for object-level editing can be labor-intensive, as both the audio track and the visual content must be edited to obtain coherent before/after pairs for each sounding video. 
Modality differences further complicate automatic dataset construction. Open-domain audio source separation is difficult when the number and types of sources are unknown~\cite{wisdom2020whatsfussfreeuniversal}, even though mixtures are easy to synthesize by linearly summing isolated tracks~\cite{oppenheim1999dtsp}. For vision, adding objects is harder than removing them, since removal can be guided by an object mask and inpainting, whereas realistic insertion must also respect scene geometry, lighting, and occlusion~\cite{wasserman2025paintinpaintlearningadd}.

\begin{figure}
\centering
  \includegraphics[width=\linewidth]{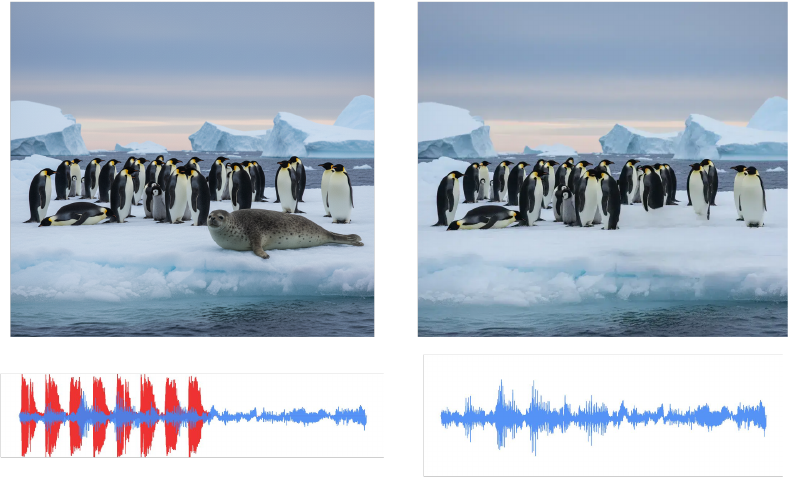}
  \caption{{\bf Object-Level Audiovisual Joint Removal.}
     During an Antarctic shoot, a penguin-colony scene is disrupted when a seal rushes into the frame and loudly calls. To avoid a costly reshoot, post-production must jointly remove the seal and its sound while maintaining the audiovisual synchronization of the edited clip.}
  \label{fig:teaser}
\end{figure}

To address the data and modeling challenges in joint audio-visual editing, we introduce \textsc{SAVEBench}, a paired audiovisual dataset with text and mask conditions, generated by a novel and scalable synthetic pipeline. With this dataset, we train the \textbf{Schr\"{o}dinger Audio-Visual Editor (SAVE)}, an end-to-end flow matching-based model for object-level joint removal in audiovisual content, incorporating a Schr\"{o}dinger Bridge~\cite{Schrodinger1932,DeBortoli2021DiffusionSchrodingerBridge} to directly transport source audiovisual content to target audiovisual content.
% \liming{Should this paragraph go into the contributions?}
We conduct extensive experiments evaluating removal success, perceptual quality, and audiovisual synchronization. The results show that \textsc{SAVE} achieves object removal performance comparable to pairwise combinations of state-of-the-art audio and video editors, while delivering higher perceptual quality, stronger temporal synchronization, and better semantic audiovisual alignment. Our demo page can be found here.\footnote{https://wx83.github.io/SAVE-Audiovisual-Joint-Removal/}

Our contributions are as follows:
\begin{itemize}
\item We introduce a synthetic pipeline that generates paired audiovisual examples with masks and text for object-level audiovisual removal.
\item We train a joint editor, \textbf{Schr\"{o}dinger Audio-Visual Editor (SAVE)}, on paired data to remove the same target object from audio and video with user instructions.
\end{itemize}
\section{Related Work}
\label{sec:formatting}

\subsection{Object Removal Datasets}
% We compare existing datasets that can support editing in Table~\ref{tab:object_removal_datasets}. We are the first multimodal paired dataset to facilitate audiovisual joint editing. Existing object-level removal datasets are usually in single modality. 
% \weihantodo{Potential Downside, one modality, no pair, total hours limited}
% For example, DAVIS, YouTube-VOS, MOSE, LVOS, ROVI, VPBench are for video editing only. Existing audio removal dataset only remove some audio effect like distroartion or remove speech, our dataset can remove object from audio track We show a detailed comparison in Section~\ref{sec:comparison_existing_dataset}. 
% % \weihantodo{If space permits, it would be better to briefly describe each existing dataset to clarify their limitations.}

Existing object‐removal resources are largely single–modality. Video inpainting datasets such as DAVIS~\cite{Perazzi2016DAVIS}, YouTube-VOS~\cite{ding2023mosenewdatasetvideo}, LVOS~\cite{hong2023lvosbenchmarklongtermvideo}, and VPBench~\cite{bian2025videopainteranylengthvideoinpainting} emphasize segmentation masks or evaluation protocols rather than releasing paired input–output edits or instruction grounding. ROVI~\cite{liang2024language} introduces instruction‐driven visual inpainting results, but focuses on visual edits only and does not include aligned audio. On the audio side, LibriMix~\cite{cosentino2020librimixopensourcedatasetgeneralizable} and VoiceBank\textendash DEMAND~\cite{Valentini2017VoiceBankDEMAND} target speech separation or enhancement and are audio‐only, not paired with video, and not aligned to specific visual objects. In contrast, our dataset supplies paired, instruction‐conditioned, object‐aligned audiovisual edits, enabling a rigorous evaluation of joint editing task. 

\subsection{Single-Modality Editing}

% \ken{Prior work on audio removal typically conditions on text \cite{wang2023audit, manor2024zeroshotunsupervisedtextbasedaudio} or on audio cues \cite{cheng2025audio, liang2025audiomorphix}. However, they often ignore the corresponding visual cues provided in the video. To overcome these limitations, we propose a joint audio-visual model that shares the same backbone, which takes into account the visual context in the video to guide the audio removal process.}

% They try to remove 
% Previous work in audio editing include

% ZEro-shot UnSupervized (ZEUS) editing \cite{manor2024zeroshotunsupervisedtextbasedaudio} introduces two approaches for zero-shot audio
% editing with pre-trained audio DDMs, one based on text
% guidance and the other based on semantic perturbations
% that are found in an unsupervised manner. All in unsupervised manner. 

% AUDIT is Diffusion based editing \cite{wang2023auditaudioeditingfollowing} build a new 
% takes the audio to be edited and the editing instruction as conditions and generates the edited audio as
% output
Prior work on audio and video editing largely treats the two modalities separately. For audio removal and editing, existing methods condition on text or audio cues without using aligned visual context: text-based diffusion approaches such as AUDIT~\cite{wang2023auditaudioeditingfollowing} and ZEUS~\cite{manor2024zeroshotunsupervisedtextbasedaudio} follow textual instructions, while audio-conditioned methods rely on reference audio or learned latent factors~\cite{cheng2025audio,liang2025audiomorphix}. These systems can remove or replace target sounds but ignore visual cues that help disambiguate on- and off-screen sources and enforce temporal alignment. On the visual side, video editors either adapt text-to-image or text-to-video diffusion backbones for object removal, or train on paired datasets. LGVI~\cite{wu2024languagedrivenvideoinpaintingmultimodal}, VideoPainter~\cite{bian2025videopainteranylengthvideoinpainting}, and VACE~\cite{jiang2025vaceallinonevideocreation} adapt text-to-video or video-diffusion backbones for object-level editing by conditioning on prompts and structured controls such as masks and text, while paired-supervision methods such as ROVI~\cite{wu2024languagedrivenvideoinpaintingmultimodal} learn directly from before/after examples. As shown in InstructPix2Pix~\cite{brooks2023instructpix2pixlearningfollowimage}, natural language instructions can serve as an expressive editing interface, but most video editors with text-to-video backbone rely on dense output captions. We adopt the paired-data paradigm and extend it to synchronized, object-level audiovisual edits with user-specified instructions.

\subsection{Audiovisual Joint Generation}

Earlier audio–visual work primarily focuses on generating a single modality conditioned on the other \cite{sung2023sound, li2022learning, iashin2021taming, pascual2024masked, su2023physics, wang2024v2a, yang2024draw, liang2024foundations}. More recent efforts shift toward joint generation or editing of an audio–video pair, typically following two paradigms. In cascaded approaches, one generates video first and then synthesizes audio conditioned on it \cite{comunita2024syncfusion, hu2024video, jeong2025read}, while the other generates audio first and then synthesizes video conditioned on the sound \cite{jeong2023power, yariv2024diverse, zhang2024audio}. This design is simple and modular but can accumulate errors and drift in cross-modal coherence. In non-cascaded approaches, audio and video are generated jointly with objectives for synchronization, semantic consistency, and temporal alignment, yielding tighter coupling; recent work shows strong correspondence and spatiotemporal alignment \cite{zhao2025uniform, liu2024syncflow, tang2023any}.
However, most joint editing methods rely on diffusion inversion~\cite{liang2024language,lin2025zero} and text-to-video backbones, which typically restrict outputs to low frame rates~\cite{henschel2025streamingt2vconsistentdynamicextendable}. In this work, we synthesize audiovisual pairs and train on generated pairs directly to perform joint audiovisual removal.

\section{SAVEBench Dataset}
Our first contribution is a new algorithm to synthetically generate large amounts of high-quality audiovisual paired data for object-level removal\footnote{We provide a detailed scalability analysis and a human evaluation of the constructed dataset in the supplementary materials in Appendix~\ref{sec:dataset_details}.}. Curating such data is labor-intensive because both the audio and visual components must be removed from each sounding video while preserving the original synchronization between modalities, which requires tightly coupled edits to audio and video. This challenge is further compounded by modality-specific difficulties: in audio, open-domain separation is fundamentally difficult~\cite{wisdom2020whatsfussfreeuniversal}, whereas mixing is well modeled~\cite{oppenheim1999dtsp}, and in vision, adding objects is harder than removing them because removal can be guided by a mask~\cite{wasserman2025paintinpaintlearningadd}. Therefore, we introduce a novel synthetic dataset pipeline that automatically detects sounding objects, synthesizes clean object-centric audio with video-to-audio model, localizes and inpaints the target visuals, and produces matched audiovisual pairs.

\subsection{Data Construction Pipeline}\label{sec:dataset_construction_pipeline}

\paragraph{Data Collection}
We collect video samples from AudioSet~\cite{gemmeke2017audioset} and KlingFoley~\cite{wang2025klingfoleymultimodaldiffusiontransformer} to construct paired samples, in each of which a chosen target object is removed from both the video and its audio. Due to resource constraints, we construct paired data using only the first 10 seconds of each video. In total, we process 50 hours video from Klingfoley a subset of audioset with around 535 hours.  

\paragraph{Target-Object Identification}
We use \textsc{Qwen-VL}~\cite{Qwen-VL} to identify objects likely to produce sound in each video and select the two most frequent across the clip. Specifically, we sample frames at 1 fps, query \textsc{Qwen-VL} for the sounding object in each frame, and then tally the two most common sounding objects over the entire video.

\paragraph{Audio Pair Construction}
For each object \(O_i \in \{O_1,\ldots,O_N\}\), we use the video-to-audio model \textsc{MMAudio}~\cite{cheng2025taming} to synthesize an object-specific track \(a_i\), conditioning on the object name and the original video to ensure audio–visual alignment and synchronization. We then apply \textsc{Qwen-Audio}~\cite{Qwen-Audio} to verify that \(a_i\) is a \emph{clean} rendition\footnote{We define a clean track as one that contains only the intended target sound, with no other prominent sources, minimal background noise, and no noticeable timing drift.}. We list prompts in our supplementary materials. Given verified tracks \(\{a_j\}_{j=1}^{N}\), we construct the mixtures as
\begin{equation}
A^{(i)}_{\setminus O_i} = \sum_{j \neq i} a_j, \qquad
A^{(i)} = A^{(i)}_{\setminus O_i} + a_i,
\end{equation}
where \(A^{(i)}_{\setminus O_i}\) is the mix with \(O_i\) removed and \(A^{(i)}\) is the full \(N\)-object mix.

\paragraph{Visual Pair Construction}
For each target object \(O_i \in \{O_1,\ldots,O_N\}\), we construct a visual pair \(\big(V^{(i)}_{\setminus O_i},\, V^{(i)}\big)\), where \(V^{(i)}\) is the original clip and \(V^{(i)}_{\setminus O_i}\) is the same clip with only \(O_i\) removed. We first localize \(O_i\) with GroundingDINO~\cite{liu2023grounding} to obtain a bounding box \(B_i\) and generate a segmentation mask \(M^{(i)}\) with SAM2~\cite{ravi2024sam2segmentimages} using the bounding box. We then apply a video inpainting model~\cite{yu2023inpaint} to \(V^{(i)}\) using \(M^{(i)}\) to erase the masked region, yielding \(V^{(i)}_{\setminus O_i}\).

\paragraph{Audiovisual Pairs}
For each object \(O_i\), we pair audio and video to create two aligned examples. The with-object pair is \(\big(V^{(i)},\, A^{(i)}\big)\) and the without-object pair is \(\big(V^{(i)}_{\setminus O_i},\, A^{(i)}_{\setminus O_i}\big)\). Because each \(a_i\) is synthesized by a video-to-audio model conditioned on the frames, the audio is frame-synchronous with the video, and generated pairs are inherently audio–visual aligned and synchronized.

\subsection{Data Statistics}
\begin{figure}
    \centering
    \includegraphics[width=0.9\linewidth]{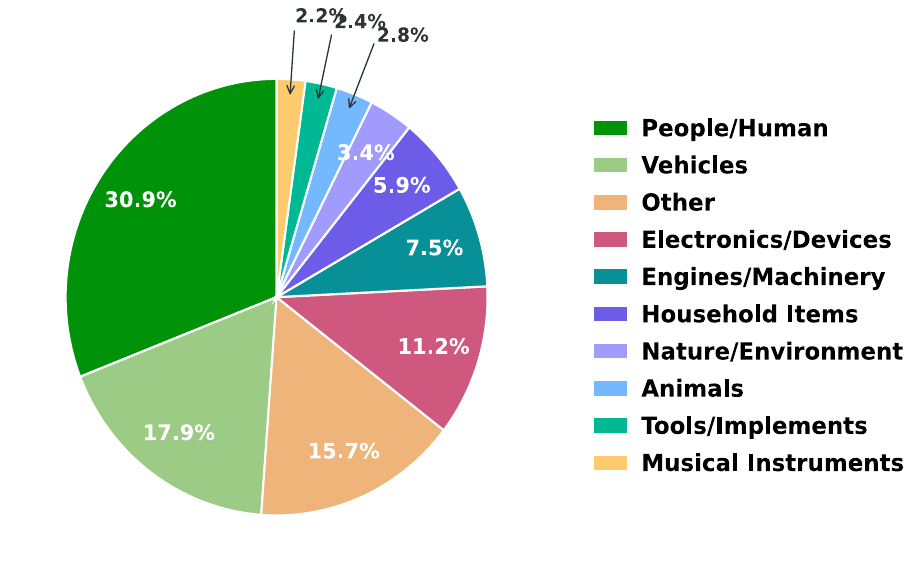}
    \vspace{-6mm}
    \caption{\textbf{Distribution of target objects.} We categorize target objects for removal into 10 categories and plot their distribution.}
    \label{fig:object_distribution}
\end{figure}

With our constructed dataset generation pipeline, we present object-level annotations with text and mask for our dataset, comprising 17{,}306 paired examples. Each instance is labeled with an object name and accompanied by a mask of the target object, enabling precise evaluation of removal and reconstruction.
% As shown in Fig.~\ref{fig:object_distribution}(b), over 95\% of the original videos are longer than 100 seconds. 
In Fig.~\ref{fig:object_distribution}, we show the distribution of detected object names categorized into 10 groups\footnote{We provide a detailed category division in our supplementary materials. Note that we use the original object names as annotations in Appendix~\ref{sec:dataset_details}.}. 

\begin{figure*}
    \centering
    \includegraphics[width=0.95\linewidth]{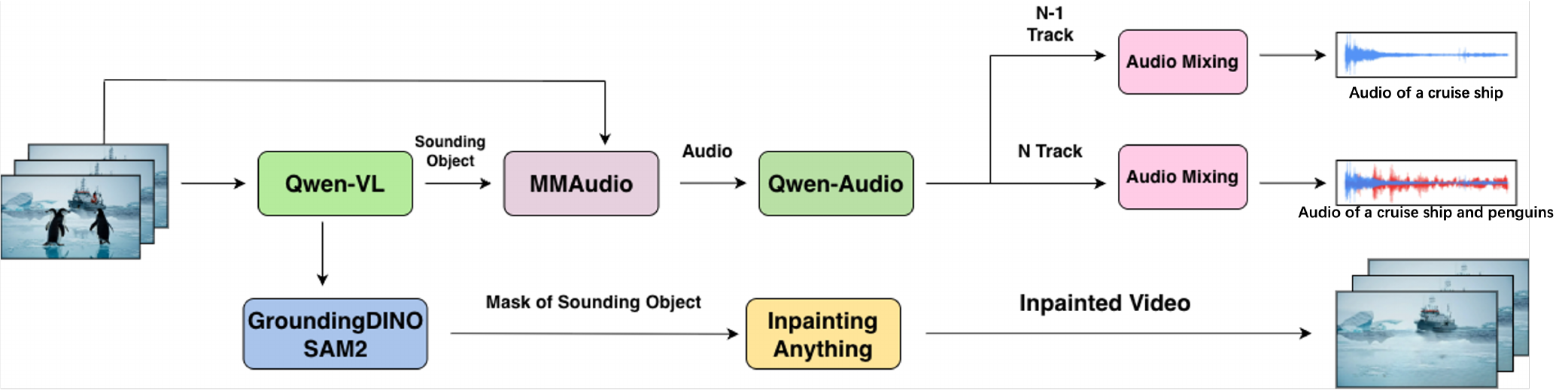}
\caption{\textbf{Dataset Pipeline Overview.} We show our synthetic pair-generation pipelines. For audio, we use Qwen-VL~\cite{Qwen-VL} to enumerate sounding objects (Penguins and Cruise), then synthesize an object-centric track for each with MMAudio~\cite{cheng2024taming}, and keep only tracks validated as clean by Qwen-Audio~\cite{Qwen-Audio}. We form \(N\!-\!1\) by mixing all retained tracks except one (Audio of a cruise ship), and \(N\) by reintroducing the held-out track (Audio of a cruise ship and penguins). For visual, we obtain object boxes with GroundingDINO~\cite{liu2023grounding} and segmentation masks with SAM2~\cite{ravi2024sam2segmentimages}. With the segmentation masks, we remove the corresponding objects (Penguins) with Inpaint-Anything~\cite{yu2023inpaint} on each frame of the video.}

    \label{fig:dataset_distribuion}
\end{figure*}

% \begin{figure}
%     \centering
%     \includegraphics[width=1\linewidth]{figures/CVPR2026.png}
%     \caption{Dataset Construction Pipeline. We first identify target sounding objects with a vision–language model, then we synthesize object-conditioned audio with visuals. Thirdly, we form clean/noisy audio pairs by controlled mixing. Then we inpaint the video to remove the target object while preserving audiovisual synchronization. \paul{font size way too small} }
%     \label{fig:dataset}
% \end{figure}

\section{Schr\"{o}dinger Audio-Visual Editor}

In this section, we develop \emph{Schr\"{o}dinger Audio-Visual Editor (SAVE)}, a new joint audio-visual generative model to perform removal. First, we formulate the problem of object-level audio-visual removal (AVR) in Section~\ref{subsec:formulation}, and then we describe our approach in Section~\ref{subsec:method}.

% We motivate is to use SB--> data hungry issue. 
% One challenge of audio-visual editing is \emph{synchronization}: When performing audio and video editing separately, the edited audio and video recordings are often temporally misaligned and semantically inconsistent, as shown in Section ~\ref{sec:combine_modality_edit_res}. Therefore, to achieve high-quality audio-visual editing, the model would need to model the audio and video \emph{jointly} to synchronize them properly. 

 % The challenge is the \emph{data-hungry} due to \emph{heterogeneity} between the audio and visual modalities. Although originating from the same sounding objects, audio and visual modalities often differ greatly in dynamic range and granularity and tend to capture different types of object properties. This discrepancy leads to conflicts in hyperparameter settings for different modalities within the same editing algorithm, resulting in imbalanced learning where the model under-fits in one modality and over-fits in the other, as shown in ~\ref{sec:weighted_loss}. 
 % \paul{also need cite evidence for this}
 
\subsection{Audio-visual Removal}
\label{subsec:formulation}
The audio-visual editor takes as input an \emph{audio-visual (AV) mixture}, i.e., a video recording $V_0\in \sR^{T_v\times H\times W\times 3}$ and an audio recording $A_0\in\sR^{T_a},$ and modifies them to create a new mixture $(V_1, A_1)\in \sR^{T_v\times H\times W\times 3}\times\sR^{T_a}$ based on some edit instruction $y\in\gY$. Here, our model takes both a \emph{textual prompt} and a user-provided \emph{segmentation mask} as inputs of the target object, which are standard in prior interactive editors (text-conditioned~\cite{bian2025videopainteranylengthvideoinpainting}, mask-conditioned~\cite{zhou2023propainter}). Our goal is to remove the target object. With our dataset construction pipeline in Sec.~\ref{sec:dataset_construction_pipeline}, the audio and video recordings are assumed to be \emph{temporally aligned} and contain the same set of sounding objects. However, due to the \emph{heterogeneity} between audio and visual modalities, maintaining synchronized edits across both streams is nontrivial and increases data demands. Therefore, we replace traditional conditional flow matching~\cite{lipman2023flow} with \emph{\sbfm/ (SBFM)}, which we will describe in Sec.~\ref{subsec:method}. 
Instead of generating edited mixtures from Gaussian white noise conditioned on the original mixtures, our approach directly interpolates between the original and edited mixtures and thus follows a much shorter sampling path. This eliminates the need to learn audiovisual reconstruction from pure noise during training.

\subsection{\sbfm/ for AVE}\label{subsec:method}
Our method consists of three components: a pair of uni-modal encoders to compress AV mixtures into low-dimensional vectors, an SBFM predictor for conditional generation in latent space, and a pair of uni-modal decoders to map generated latents back to sample space.

\paragraph{Feature Extraction} To remove low-level, non-semantic information from the AV mixture, we first extract compact latent vectors for the original and edited mixtures using a pair of frozen encoders $e_a:\sR^{T_a}\mapsto \sR^{T d_a},\,e_v:\sR^{T_v\times H\times W\times 3}\mapsto\sR^{Td_v}$ from a pair of pretrained audio-only variational auto-encoder (VAE) SoundCTM~\cite{saito2025soundctm} and video-only variational auto-encoder CVVAE~\cite{zhao2024cvvae}. To match the audio and video timelines, we add a lightweight temporal projector \(P_\tau:\mathbb{R}^{T_v d_v}\!\to\mathbb{R}^{T_a d_v}\) that aligns the video latent sequence to the audio temporal grid to learn audiovisual synchronization. Then, we concatenate audio feature dimension $d_a$ and video feature dimension $d_v$ and get AV mixture in latent space:
\begin{equation}\label{eq:avmixture}
\begin{aligned}
    X^a_m &:= e_a(A_m),\, X^v_i:= e_v(V_m),\\
    X_m&:=[X^a_m, P_\tau(X^v_m)]\in \sR^d,
\end{aligned}
\end{equation}
for $m\in\{0,1\}$. 

% Note we perform resampling over the two modalities to align their temporal dimension.

\paragraph{Conditions}
% We use CLAP\cite{wu2024largescalecontrastivelanguageaudiopretraining} to extract the audio emebddinsg corespoing to the text condition and use a semantic mask where we element wise times the clip embeddings on the mask. 
% \paul{this part seems important. it should be reflected in the figures, where u show the full image, and after it is multiplied by the mask, and the highlighted/masked image that comes out. these images should be part of figure 2}
% We multiply the video frames $\bm{v}$ with segmentation mask frames $\bm{v}_{sam}$ to obtain the masked video frames $\bm{v}_{masked}\in\mathbb{R}^{3\times T_v\times H\times W}$, and encode with a semantic visual encoder C-RADIOv3 \cite{heinrich2025radiov2} to obtain visual features $\bm{z}_s\in\mathbb{R}^{C_v\times T_v\times H'\times W'}$
% (\texttt{630k-audioset-best.pt}) 
To reliably edit the \emph{right} object in both audio and video, we bridge the user-specified condition to audio with a pretrained text–audio model and use a segmentation mask to indicate object presence in the visual stream. We use CLAP~\cite{wu2024largescalecontrastivelanguageaudiopretraining} to obtain a text-conditioned audio embedding \(\phi_a\), which specifies what sound to keep or remove. To localize where this condition applies in the frames, we gate the pixels with a semantic mask. Concretely, we resize video frames and SAM2~\cite{ravi2024sam2segmentimages} masks to \(128\times128\), with frames \(V_0 \in \mathbb{R}^{T_v\times H\times W\times 3}\) and masks \(M \in [0,1]^{T_v\times H\times W\times 3}\), and obtain the visual condition by
\begin{equation}
\phi_v = \Xi_{\text{vis}}(M \odot V_0),
\end{equation}
where \(\Xi_{\text{vis}}\) is the C\text{-}RADIOv3~\cite{heinrich2025radiov25improvedbaselinesagglomerative} semantic visual encoder.
We inject \(\phi_v\) via cross-attention and add \(\phi_a\) to the audio timestep embeddings, yielding the joint condition \(y = \{\phi_a, \phi_v\}\) for flow matching with \(X_0 \sim p_0(\cdot \mid y)\).

% \weihantodo{We also have two linear mlp to lower the dimension}
% and condition the model on \([\,e_a;\,\mathrm{pool}(\tilde{f}_v)\,]\). 

\paragraph{Conditional Flow Matching} 
Using the latent features, we then treat AVR as a conditional generation problem. A common approach used in uni-modal generation is \emph{conditional flow matching (CFM)}~\cite{lipman2023flow,Tong2024ImprovingCFM}. Here, we adopt a more general framework where the starting point of the flow is not restricted to be Gaussian. Intuitively, a CFM conditioned on  $y\in\gY$ learns the distribution $X_1\sim p_1(\cdot\mid y)$ by \emph{moving} samples from an initial distribution $X_0\sim p_0(\cdot\mid y)$ in a \emph{deterministic} manner. This is captured by the following ordinary differential equation (ODE):
\begin{align}
    \frac{\dd X_t}{\dd t}=v^{\theta}_t(X_t,y),\,X_0\sim p_0(\cdot\mid y),\,t\in[0,1],
\end{align}
where the velocity field $v_t^{\theta},\,t\in[0,1]$, or \emph{flow}, specifies how the samples at each location should be moved at a given time with trainable parameters $\theta$, which are learned from \emph{paired} samples $(X_0, X_1, y)$. Directly estimating the true flow $u_t(X_t,y)$ is often not possible for complex data such as AV mixtures since it requires sampling from $p_t$ and oracle access to $u_t$. Nevertheless, we can often find a simple, closed-form solution for the \emph{conditional} flow $u_t(X_t|X_0,X_1)$. For instance, for the classical CFM, $p_0$ is a standard Gaussian, and the conditional flow is simply
\begin{align}\label{eq:cfm_cond_flow}
    u_t^{\mathrm{CFM}}(X_t|X_0,X_1)=X_1-X_0,\,\forall t\in[0,1]
\end{align}
for $X_t=tX_1+(1-t)X_0=:\mu_t$ and any final distribution $p_1$~\cite{lipman2023flow}. Note that in this case $p_{t|0,1}=\delta_{\mu_t}$ is a delta function and $p_{01}(x_0,x_1\mid y)=p_0(x_0\mid y)p_1(x_1\mid y)$ is an independent measure. With a tractable conditional flow,
%with instantaneous target velocity
%\[
%u_t=\dot{\alpha}(t)(X_1-X_0)+\dot{\sigma}(t)\varepsilon.
%\]
CFM then trains $v_\theta$ by a simple regression loss:
\begin{align}\label{eq:cfm}
\Ls_{\mathrm{CFM}}=\mathbb{E}_{t,p_{01},p_{t|01}}\,\big\|\,v^{\theta}_t(X_t,y)-u_t^{\mathrm{CFM}}(X_t|X_0,X_1)\,\big\|_2^2,
\end{align}
which has the same optimal solution as directly matching with the true flow~\cite{lipman2023flow}.

% Use the noisy linear coupling
% \[
% X_t=\alpha(t)X_1+\big(1-\alpha(t)\big)X_0+\sigma(t)\varepsilon,\quad \varepsilon\!\sim\!\mathcal{N}(0,I),
% \]

% \paul{need to explain why we are doing all this, not just state it. also clarify what is existing, what is new that u modified}

% Let $X_1\sim p_1(\cdot\mid y)$ be data conditioned on $c$, and $X_0\!\sim p_0$ a simple prior.
% We learn a conditional vector field $v_\theta(x,t,c)$ and sample by integrating
\begin{figure*}
    \centering
    \vspace{-6mm}
    \includegraphics[width=1\linewidth]{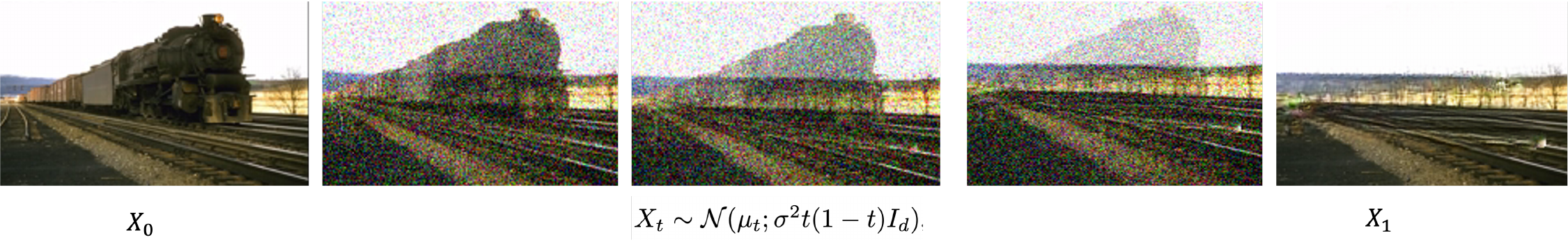}
    \caption{{\bf \sbfm/.} Given source latent $X_0$ and latent $X_1$, the state variable $X_t$ is sampled from  $\gN(\mu_t;\sigma^2t(1-t)I_d)$. Notice that in traditional conditional flow matching, we sample $X_0$ as a Gaussian. }
    \label{fig:sb}
\end{figure*}
\paragraph{\sbfm/} 
% \paul{again, call it audiovisual schrodinger or something.. and explain what is new about it.. what did u have to change to apply it for audio video data and to synchronize it..} 
Instead of choosing $p_0$ as a Gaussian, we set it to be the distribution of the latent features of the original AV mixtures in Eq.~\ref{eq:avmixture}. To derive the correct flow between two arbitrary distributions $p_0$ and $p_1$, we use a \emph{Schr\"{o}dinger bridge (SB)}~\cite{Schrodinger1932}, which tries to find the \emph{most efficient} sampling path between distributions by solving a stochastic, dynamic optimal transport problem~\cite{Chen2014OnTR,DeBortoli2021DiffusionSchrodingerBridge}. 
% The solution to SB can be expressed by the following forward and backward stochastic differential equations (SDEs)~\cite{DeBortoli2021DiffusionSchrodingerBridge,chen2022likelihood}:
% \begin{align}\label{eq:sb_sde}
%     \dd X_t &= [f_t+\beta_t\nabla\log\Psi_t]\dd t+\sqrt{\beta_t}\dd W_t,\,X_0\sim p_0,\\
%     \dd X_t &= [f_t+\beta_t\nabla\log\hat{\Psi}_t]\dd t+\sqrt{\beta_t}\dd \overline{W}_t,X_1\sim p_1,
% \end{align}
% where $f_t(X_t)\in \sR^d$ controls the drift, $\beta_t\in\sR$ is the diffusion, and $W_t,\overline{W}_t\in\sR^d$ are standard forward and backward Wiener processes, respectively. The additional drift terms $\nabla \log \Psi_t(X_t,y)$ and $\nabla\log \hat{\Psi}_t(X_t,y)$ are score functions satisfying the following coupled partial differential equations (PDE):
% \begin{equation}
% \begin{aligned}
%     &\begin{cases}
%         \frac{\partial\Psi}{\partial t}=-\nabla\Psi^\top f-\frac{1}{2}\beta\Delta\Psi\\
%         \frac{\partial\hat{\Psi}}{\partial t}=-\nabla\cdot(\hat{\Psi} f)+\frac{1}{2}\beta\Delta\hat{\Psi}
%     \end{cases},\Psi_0\hat{\Psi}_0=p_0,\Psi_1\hat{\Psi}_1=p_1.
% \end{aligned}
% \end{equation}
% when $f_t\equiv 0$ and $\beta_t\equiv \sigma^2,$ 
As shown in our supplementary materials and \cite{Tong2024ImprovingCFM}, the conditional flow of SB has the following closed-form expression: 
%\paragraph{Learning Schr\"{o}dinger's bridge (SB) with paired data}
%\begin{align}\label{eq:sb_prob}
%    X_t|x_0,x_1\sim \mathcal{N}\left(tx_1+(1-t)x_0;\sigma^2t(1-t)\right).
%\end{align}
\begin{align}\label{eq:sb_cond_flow}
    u_t(X_t|X_0,X_1)=u_t^{\mathrm{CFM}}+\frac{1-2t}{2t(1-t)}(X_t-\mu_t),
\end{align}
where $X_t\sim \gN(\mu_t;\sigma^2t(1-t)I_d)$, which is pinched on $X_0$ and $X_1$ at the starting and end times, respectively. We show a schematic in Fig.~\ref{fig:sb}. 
% Note that the second term in Eq.~\ref{eq:sb_cond_flow} is a correction term to the sampling path in the classical CFM to account for the stochasticity of $X_t$ due to $\sigma>0$, since we observe that a nonzero $\sigma$ leads to better generalization performance for AVE. 
A naive approach will be to train the SBFM using Eq.~\ref{eq:cfm} with Eq.~\ref{eq:sb_cond_flow} as the conditional flow. 
However, the heterogeneity between the two modalities makes it necessary to \emph{reweight} the loss for the audio and video components \(v_t^{\theta} := [v_t^{\theta,a}, v_t^{\theta,v}]\) of the flow:
\begin{multline}\label{eq:loss_curve}
    \Ls_{\mathrm{SAVE}}=\E_{t,p_{01},p_{t|01}}\big(\|v_t^{\theta,a}(X_t,y)-u_t^a(X_t^a \mid X_0^a,X_1^a)\|^2 \\
    +\; \lambda\|v_t^{\theta,v}(X_t,y)-u_t^v(X_t^v \mid X_0^v,X_1^v)\|^2\big),
\end{multline}
where supervision is applied per modality, with \(X_t\) including \(X_t^a\) and \(X_t^v\), \(\lambda > 0\) a hyperparameter that balances the audio and video losses, and \([u_t^a, u_t^v] := u_t\) denoting the audio and video components of the conditional flow\footnote{We provide a detailed math derivation in Appendix~\ref{sec:details_sb_math}}.

\paragraph{Uni-modal Decoders}
As shown in Sec.~\ref{sec:disentangle}, simply splitting audio and video along the feature dimension makes it hard to disentangle the audiovisual mixture. Therefore, we add an audio head and a video head to separate the learned audiovisual velocity \(v_t\) from the DiT backbone into modality-specific velocities \(v_t^a\) and \(v_t^v\). We use a small stack of DiT blocks for the audio head and another stack for the video head. During sampling, we integrate the ODE separately using \(v_t^a\) and \(v_t^v\) to obtain \(X_t^a\) and \(X_t^v\), then decode \(X_t^a\) and \(X_t^v\) with the audio and video decoders, reusing the same audio and video VAEs used for feature extraction.

% /mnt/Datasets/audioset_kling_processed/a/u/audioset_eval_FRoMl49q5gc_160.000
\begin{figure*}
    \centering
    \includegraphics[width=0.9\linewidth]{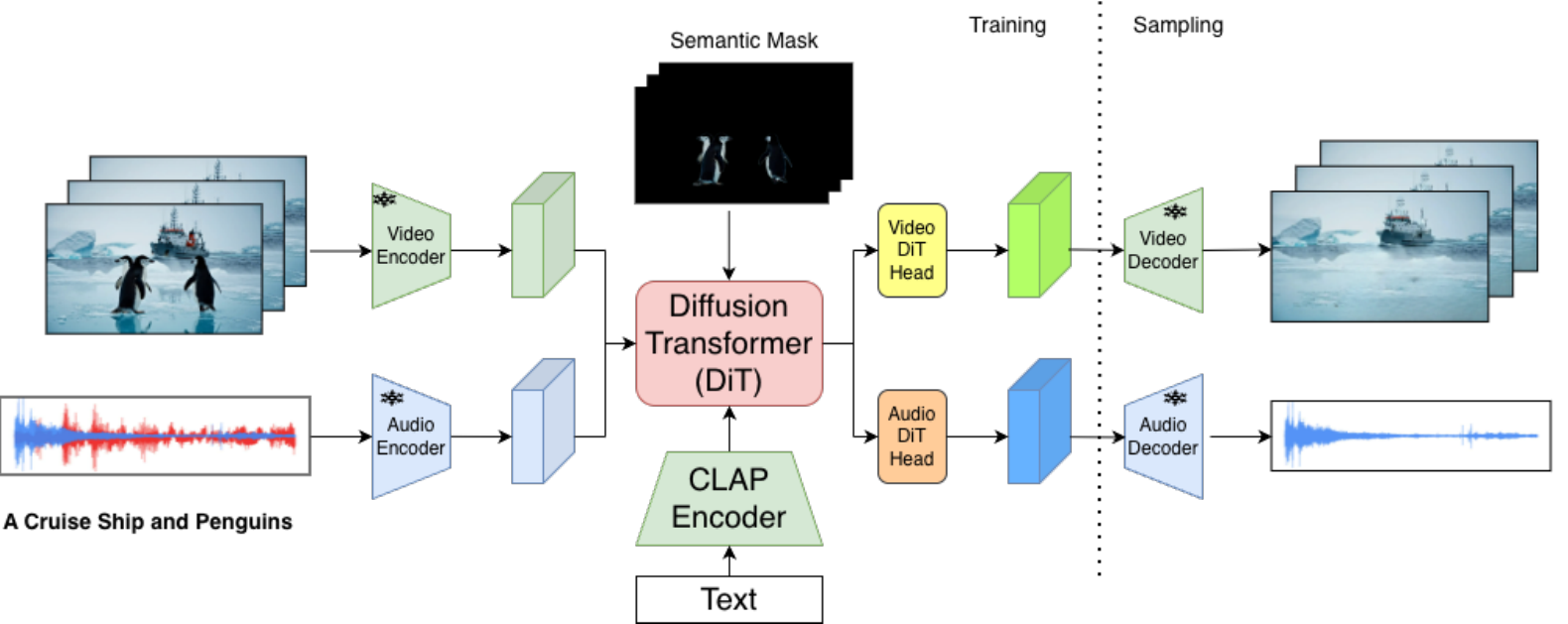}
    \vspace{-2mm}
    \caption{{\bf Model architecture.} Given an input audio and video pair, our goal is to remove a target sounding object from both audio and video. We employ pre-trained audio and video VAE encoders to encode audio and video into their respective latent representations. We then feed them into our SAVE model together with the semantic mask and CLAP embeddings that guide the removal process for the selected object. Finally, we use a pre-trained audio VAE decoder and a pre-trained video VAE decoder to reconstruct the waveform and video frames from the latent space, respectively. Note that the VAE encoders for the target audio and video are not used at test time. }
    \label{fig:model_achitecture}
\end{figure*}

\section{Experiment Setup}

\subsection{Implementation Details}
\paragraph{Datasets and Configuration}
 We train our dataset on the constructed SAVEBench. We keep only videos longer than 8 seconds and crop them to the first 8 seconds due to resource constraints. In total, the dataset includes 15,575 training pairs, 865 validation pairs, and 866 test pairs. 
 
\paragraph{Configuration}
We sample audio at 44.1 kHz and resize video frames to 128×128 for audio and video VAEs. The backbone DiT has hidden dimension 2048 with 16 layers and 16 attention heads, and we attach two small DiTs with 4 layers and 4 attention heads each as audio and video heads to disentangle the mixture. For sampling, we use the Euler method to solve the ODE with 30 steps for both audio and video latents. We train the model on four H100 GPUs using AdamW~\cite{loshchilov2019decoupledweightdecayregularization} with $\beta_1 = 0.9$, $\beta_2 = 0.999$, learning rate warmed up from $10^{-5}$ to $10^{-4}$, and weight decay $10^{-4}$, and select the checkpoint with lowest validation loss at epoch 42. We set $\lambda = 3$ for the main model and provide the full training configuration in the supplementary materials.

% Wrote: /home/weihan.xu/av-cfm-two-noise-edit-mask-concat/avedit_training/train.json  (15575 items)
% Wrote: /home/weihan.xu/av-cfm-two-noise-edit-mask-concat/avedit_training/valid.json  (865 items)
% Wrote: /home/weihan.xu/av-ci thinkfm-two-noise-edit-mask-concat/avedit_training/test.json   (866 items)
% Total: 17306 (seed=42)

\subsection{Baselines}

We compare our model against two baseline types: one is cross-modal baselines, where we cross two audio editors with three video editors, and the other is a multimodal editor trained with a conditional flow matching objective\footnote{We intend to include joint audiovisual inpainting baselines~\cite{liang2024languageguidedjointaudiovisualediting,lin2025zeroshotaudiovisualeditingcrossmodal,fu2025objectaveditobjectlevelaudiovisualediting}. However, their codebases are not publicly available.}.

\subsubsection{Cross-Modal Baselines}
% We construct six cross-modal baselines by crossing two audio editors with three video editors, yielding all 6 pairings. For audio-editing, we assess: (i) \textbf{ZEUS~\cite{manor2024zeroshotunsupervisedtextbasedaudio} } which uses DDPM inversion with pretrained diffusion models for zero-shot, unsupervised, text-based audio editing. We directly run inference on our test set with their checkpoint. (ii)\textbf{AUDIT~\cite{wang2023auditaudioeditingfollowing}} is an instruction-guided audio editing model based on latent diffusion, We retrain AUDIT on SAVEBench and use it as a audio removal baseline. For video editing, we access:(i)\textbf{LGVI\cite{wu2024languagedrivenvideoinpaintingmultimodal}}, which is a diffusion-based language-driven video pinatering that leverage Multimodal large laague odel to understand and excete inpaint requres 
% (ii)\textbf{VideoPainter~\cite{bian2025videopainteranylengthvideoinpainting} is a dual-branch video inpainting framework using text-to-video backbone based which expects a descriptive prompt, we use a video without the \texttt{\#Object} tag for inference}
% (iii)\textbf{VACE~\cite{jiang2025vaceallinonevideocreation}} is and expects a descriptive prompt, we use a video without the \texttt{\#Object} tag.

% \footnote{LGVI composes publicly released checkpoints into an inference-only pipeline. We intend to retrain VACE. However, the training pipeline is not publicly available. For VideoPainter, retraining would require dataset-specific captions and is out of scope for this baseline.}
 
We construct six cross-modal baselines by combining two audio editors with three video editors, yielding six pairings. For audio editing, we consider two baselines: (i) \textbf{ZEUS}~\cite{manor2024zeroshotunsupervisedtextbasedaudio}, which uses DDPM inversion with pretrained diffusion models for zero-shot, unsupervised, text-based audio editing, and we directly run inference on our test set with their released checkpoint, and (ii) \textbf{AUDIT}~\cite{wang2023auditaudioeditingfollowing}, an instruction-guided audio editing model based on latent diffusion, which we retrain on SAVEBench and use as an audio removal baseline.\footnote{Details can be found in Appendix~\ref{sec:audit}.} For video editing, we consider three baselines: (i) \textbf{LGVI}~\cite{wu2024languagedrivenvideoinpaintingmultimodal}, a diffusion-based, language-driven video inpainting model that leverages a multimodal large language model to understand and execute inpainting requests, (ii) \textbf{VideoPainter}~\cite{bian2025videopainteranylengthvideoinpainting}, a dual-branch video inpainting framework built on a text-to-video backbone that expects a prompt, for which we use the caption \texttt{A video without \{object\}.} at inference time, and (iii) \textbf{VACE}~\cite{jiang2025vaceallinonevideocreation}, an all-in-one video creation and editing framework built on pretrained text-to-video backbones that also expect a prompt, where we similarly use the caption \texttt{A video without \{object\}.} at inference time\footnote{LGVI composes publicly released checkpoints into an inference-only pipeline. We intended to retrain VACE, but its training pipeline is not publicly available. For VideoPainter, retraining would require dataset-specific captions and is out of scope for this baseline.}.

\subsubsection{Multimodal Conditional Flow Matching}
We train a multimodal conditional flow matching (MCFM) model with the same hardware setup and controls as SAVE. However, instead of using a Schrödinger bridge, we directly transport random noise to the target data distribution using the conditional flow matching objective. In particular, we sample an initial Gaussian audio noise and an initial Gaussian video noise. We train this model on the same amount of data as SAVE, with hidden dimension 2048, comprising 16 layers and 16 attention heads, and train it for 48 hours.
% We do inference with the traditional flow matching in the same step as \sbfm/. We add separate audio noise and video noise to both modalities. 

\subsection{Evaluation Metrics}
We evaluate our model by measuring whether it removes the target object from both visual and audio content while leaving all other content unchanged, and whether the audiovisual output maintains temporal synchronization and semantic alignment\footnote{We also provide an additional subjective evaluation in Appendix~\ref{sec:subjective_test}}. 

\paragraph{Target Object Removal}
We evaluate removal quality by comparing our model outputs with the ground-truth edited pairs in \textsc{SAVEBench}.
For video removal, following \cite{wu2024languagedrivenvideoinpaintingmultimodal, bian2025videopainteranylengthvideoinpainting}, we report PSNR~\cite{PSNRSSIM}, which measures pixel-level distortion, SSIM~\cite{PSNRSSIM, SSIM}, which measures structural similarity, LPIPS~\cite{zhang2018unreasonableeffectivenessdeepfeatures}, which reflects perceptual distance in a learned feature space. We also report FVID~\cite{Wang2018Vid2Vid}, which measures perceptual quality relative to the real videos, following \cite{bian2025videopainteranylengthvideoinpainting}. For a fair comparison, we resize all frames to \(128\times128\) and set the frame rate to 8 fps.
For audio removal, following \cite{wang2023auditaudioeditingfollowing, liu2024audioldm2learningholistic}, we report Log-Spectral Distance(LSD), Fréchet Audio Distance(FAD), Inception Score(IS), and Kullback–Leibler divergence(KL). LSD quantifies the distance between the spectrograms of outputs and targets, FAD assesses distribution-level fidelity between generated and target samples, IS reflects the quality and diversity of generations, and KL divergence measures the discrepancy between output and target distributions. For a fair comparison across metrics, we resample all audio to 32\,kHz and use a CNN-14 backbone.

\paragraph{Audiovisual Alignment}
We assess preservation of audiovisual alignment after removal using two complementary metrics, DeSyncScore (DeSync) for temporal alignment and ImageBindScore (IBScore) for semantic correspondence. Following \cite{cheng2024taming}, DeSyncScore measures misalignment in seconds predicted by Synchformer~\cite{iashin2024synchformerefficientsynchronizationsparse} trained on AudioSet, averaged over the first 4.8 seconds and last 4.8 seconds of each audio–video pair, and captures whether sound onsets and offsets align with the corresponding visual events. ImageBindScore~\cite{cheng2024taming} measures semantic correspondence by extracting audio and video embeddings with ImageBind~\cite{girdhar2023imagebindembeddingspacebind} and computing their cosine similarity, indicating how well the audio content matches the visual content. When the audio and video have different lengths, we truncate the longer modality to match the shorter one.

\section{Results}
\begin{figure*}
    \centering
    \includegraphics[width=1\linewidth]{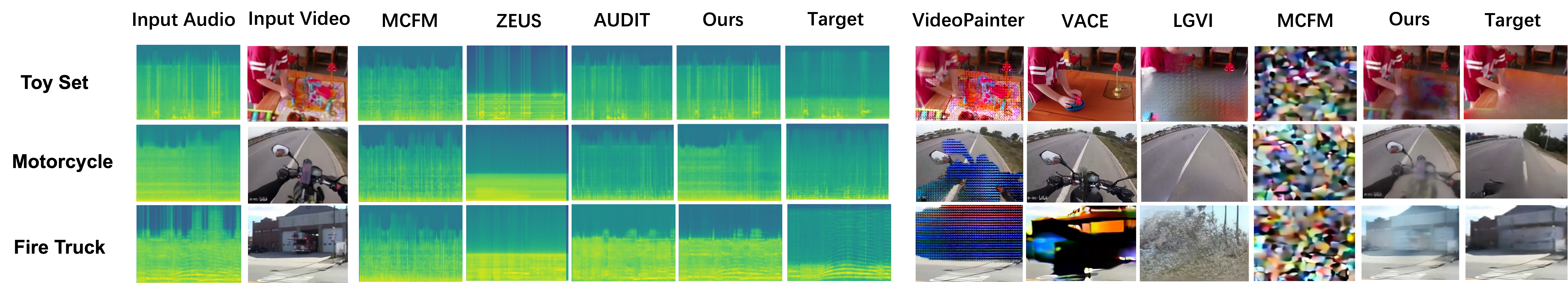}
    \caption{\textbf{Qualitative Examples.} We present three qualitative examples with a toy set, a motorcycle and a fire truck as the target object.}
    \label{fig:qualitative_example}
\end{figure*}
\subsection{Qualitative Examples}
We show some qualitative examples in Fig.~\ref{fig:qualitative_example}. Compared with visual editor baselines such as VACE\cite{jiang2025vaceallinonevideocreation}, LGVI\cite{wu2024languagedrivenvideoinpaintingmultimodal} and VideoPainter\cite{bian2025videopainteranylengthvideoinpainting}, our model can remove more pixel information of the target object while keeping the surrounding content unchanged. We also find that text-to-video visual removal baselines such as VACE and VideoPainter exhibit distortions that degrade quality, suggesting that these text-to-video backbone models rely on dense output captions and may struggle with user-specified instructions. In addition, we find that although MCFM models are trained for longer with the same number of examples and the same hardware setup as our main model, their outputs are mostly noise, indicating that they struggle with reconstructing audiovisual content. For audio editor baselines, ZEUS~\cite{manor2024zeroshotunsupervisedtextbasedaudio} removes the largest amount of audio energy. However, spectrograms show that ZEUS also suppresses additional high-frequency content than the target edited pair. Our model removes slightly less energy overall but better preserves non-target components, leading to a higher output fidelity.

\subsection{Quantitative Results}

\paragraph{Target Object Removal} We evaluate whether the target object is removed from both visual and audio content while all other content remains unchanged by comparing the outputs of our model and all baselines against the ground-truth edited pairs from \textsc{SAVEBench} in Table~\ref{tab:video_edit_performance} and Table~\ref{tab:audio_edit_performance}. For video, we assess how well each method removes the target object while preserving the rest of the scene by comparing its output to the target edited pair from \textsc{SAVEBench}. Our model achieves the best PSNR, SSIM, and LPIPS, as well as the lowest FVID, showing that it preserves fine details and global structure (PSNR, SSIM), minimizes perceptual distance (LPIPS), and produces more realistic videos (FVID) than the baselines. Although MCFM is trained longer with the same number of samples, it achieves the highest FVID and the worst reconstruction metrics. Qualitative examples further show that it even struggles to faithfully reconstruct the surrounding visual content. For audio, we similarly compare each method’s output with the ground-truth edited pair from \textsc{SAVEBench}. Our model achieves the lowest FAD, indicating the closest match to the target feature distribution, and the highest Inception Score, indicating high-quality and diverse generation. It also obtains the second-lowest KL divergence after AUDIT~\cite{wang2023audit}, showing that its predicted distribution is close to the ground truth. Additionally, it yields the highest PSNR and SSIM together with the lowest LSD, indicating the best reconstruction fidelity to the ground-truth edited audio.

\begin{table}[ht]
  \centering
  \footnotesize % changed from \scriptsize to \footnotesize
  \setlength{\tabcolsep}{5pt}
  \renewcommand{\arraystretch}{0.95}
  \begin{threeparttable}
    \begin{tabular}{lccccc}
      \toprule
      \textbf{Model} &
      \textbf{PSNR$\uparrow$} &
      \textbf{SSIM$\uparrow$} &
      \textbf{LPIPS$\downarrow$} &
      \textbf{FVID$\downarrow$} \\
      \midrule
      VACE~\cite{jiang2025vaceallinonevideocreation} &
        12.82 $\pm$ 4.03 &
        0.40 $\pm$ 0.17 &
        0.45 $\pm$ 0.16 &
        26.59 \\
      LGVI~\cite{wu2024languagedrivenvideoinpaintingmultimodal} &
        12.22 $\pm$ 3.26 &
        0.27 $\pm$ 0.17 &
        0.52 $\pm$ 0.14 &
        20.22 \\
      VideoPainter~\cite{bian2025videopainteranylengthvideoinpainting} &
        12.42 $\pm$ 3.33 &
        0.37 $\pm$ 0.20 &
        0.49 $\pm$ 0.17 &
        26.46 \\
      MCFM &
        10.03 $\pm$ 1.21 &
        0.10 $\pm$ 0.03 &
        0.05 $\pm$ 0.00 &
        139.51 \\
      Ours &
        \textbf{20.95} $\pm$ 5.34 &
        \textbf{0.69} $\pm$ 0.16 &
        \textbf{0.02} $\pm$ 0.01 &
        \textbf{13.88} \\
      \bottomrule
    \end{tabular}
    \caption{Quantitative Result for Visual Target Object Removal}
    \label{tab:video_edit_performance}
  \end{threeparttable}
\end{table}

\begin{table}[ht]
  \centering
  \scriptsize
  \setlength{\tabcolsep}{3pt}        % default ~6pt; smaller = tighter columns
  \renewcommand{\arraystretch}{0.95} % tighter rows
  \begin{threeparttable}
    \begin{tabular}{lcccccc}
      \toprule
      \textbf{Model} &
      \textbf{FAD$\downarrow$} &
      \textbf{IS$\uparrow$} &
      \textbf{KL Div$\downarrow$} &
      \textbf{PSNR (dB)$\uparrow$} &
      \textbf{SSIM (dB)$\uparrow$} &
      \textbf{LSD$\downarrow$}\\
      \midrule
      ZEUS~\cite{manor2024zeroshotunsupervisedtextbasedaudio} &
        2.69 &
        4.02 $\pm$ 0.27 &
        4.61 &
        14.60 $\pm$ 2.66 &
        0.33 $\pm$ 0.13 &
        6.28 \\
      AUDIT~\cite{wang2023audit} &
        2.85 &
        1.01 $\pm$ 0.00 &
        \textbf{0.37} &
        17.15 $\pm$ 3.64 &
        0.40 $\pm$ 0.15 &
        1.51 \\
      MCFM &
        15.82 &
        1.45 $\pm$ 0.03 &
        9.06 &
        14.54 $\pm$ 2.14 &
        0.20 $\pm$ 0.06 &
        1.77 \\
      Ours &
        \textbf{0.69} &
        \textbf{5.53} $\pm$ 0.47 &
        2.10 &
        \textbf{18.15} $\pm$ 4.23 &
        \textbf{0.51} $\pm$ 0.16 &
        \textbf{1.44} \\
      \bottomrule
    \end{tabular}
    \caption{Quantitative Result for Audio Target Object Removal}
    \label{tab:audio_edit_performance}
  \end{threeparttable}
\end{table}

\paragraph{Audiovisual Alignment}\label{sec:MCFM_result}
In this experiment, we compare audiovisual temporal alignment and semantic correspondence between our model and the baselines. As shown in Table~\ref{tab:audiovisual_sync}, our model achieves the lowest DeSyncScore, followed by MCFM, indicating the least temporal misalignment. This suggests that simply combining an audio editor with a video editor yields greater temporal misalignment. Our model also attains the highest ImageBindScore, indicating the best audiovisual semantic alignment among the baselines. The combination of the lowest DeSyncScore and highest ImageBindScore demonstrates the necessity of joint audiovisual learning for the removal task. Notably, MCFM shows the weakest audiovisual semantic correspondence, with the lowest ImageBindScore, because its outputs are mostly noise, as shown in Fig.~\ref{fig:qualitative_example}. 

\begin{table}
  \centering
  \scriptsize
  \setlength{\tabcolsep}{4pt}        % default ~6pt
  \renewcommand{\arraystretch}{0.95} % tighter rows
  \begin{tabular}{l l c c}
    \toprule
    \textbf{Video Model} & \textbf{Audio Model} & \textbf{DeSync$\downarrow$} & \textbf{IBScore$\uparrow$} \\
    \midrule
    VACE~\cite{jiang2025vaceallinonevideocreation} & ZEUS~\cite{manor2024zeroshotunsupervisedtextbasedaudio} & 1.20 & 0.15 \\
    VACE & AUDIT~\cite{wang2023audit} & 1.20 & 0.16 \\
    LGVI & ZEUS  & 1.30 & 0.12 \\
    LGVI & AUDIT & 1.31 & 0.12 \\
    VideoPainter~\cite{bian2025videopainteranylengthvideoinpainting} & ZEUS & 1.17 & 0.16 \\
    VideoPainter & AUDIT & 1.24 & 0.14 \\
    MCFM & MCFM & 1.08 & 0.01 \\
    Ours & Ours & \textbf{0.81} & \textbf{0.20} \\
    \bottomrule
  \end{tabular}
  \caption{Audiovisual alignment}
  \label{tab:audiovisual_sync}
\end{table}
% \subsection{Inference Efficiency}
% \begin{table}[ht]
%   \centering
%   \scriptsize
%   \setlength{\tabcolsep}{4pt}        % default ~6pt; smaller = tighter columns
%   \renewcommand{\arraystretch}{0.95} % tighter rows
%   \label{tab:placeholder}
%   \begin{tabular}{l l r r}
%     \toprule
%     \textbf{Video Model} & \textbf{Audio Model} & \textbf{FLOPs (G)} & \textbf{Avg. Time (s)} \\
%     \midrule
%     VACE\cite{jiang2025vaceallinonevideocreation} & ZEUS\cite{manor2024zeroshotunsupervisedtextbasedaudio} & -- & -- \\
%     VACE & AUDIT & -- & -- \\
%     LGVI & ZEUS  & -- & -- \\
%     LGVI & AUDIT & -- & -- \\
%     ProPainter & ZEUS & -- & -- \\
%     ProPainter & AUDIT & -- & -- \\
%     VideoPainter & ZEUS & -- & -- \\
%     VideoPainter & AUDIT & -- & -- \\
%     Ours & Ours & -- & -- \\
%     \bottomrule
%   \end{tabular}
%   \caption{Inference Efficiency}
%     \label{tab:audiovisual_inference}
% \end{table}

% \subsection{Ablation Study}

% We evaluate the effect of weights for audio and video in multimodal flow matching. 

% When the video is three times of video

% When the video is five times of video

% When the video is the same weight as audio
\section{Ablations}
\paragraph{Experiments of Different Loss Weight}\label{sec:weighted_loss}

In this experiment, we vary the audio-loss weight $\lambda$ in Eq.~\ref{eq:loss_curve} (1, 3, and 5) to probe heterogeneity between the audio and visual modalities. As shown in Table~\ref{tab:loss_weight_ablation_av}, setting $\lambda=3$ yields the highest PSNR/SSIM and the lowest LPIPS/FVID, indicating higher output fidelity. In Table~\ref{tab:loss_weight_ablation_sync}, we find that when we set $\lambda=3$, we obtain the lowest DeSyncScore, indicating better audiovisual temporal alignment. This further demonstrates modality differences in joint learning.

% \begin{table*}
%   \centering
%   \footnotesize
%   \setlength{\tabcolsep}{4pt}
%   \renewcommand{\arraystretch}{0.95}
%   \vspace{.5ex}
%   \begin{tabular}{lcccccccc}
%     \toprule
%     & \multicolumn{4}{c}{\textbf{Video Evaluation}} & \multicolumn{2}{c}{\textbf{Audio Evaluation}} & \multicolumn{2}{c}{\textbf{Audiovisual Synchronization}}\\
%     \cmidrule(lr){2-5}\cmidrule(lr){6-7}\cmidrule(lr){8-9}
%     \textbf{$\lambda$}
%       & \textbf{PSNR$\uparrow$} & \textbf{SSIM$\uparrow$} & \textbf{LPIPS$\downarrow$} & \textbf{FVID$\downarrow$}
%       & \textbf{FAD$\downarrow$} & \textbf{LSD$\downarrow$}
%       & \textbf{DeSyncScore$\downarrow$} & \textbf{ImageBindScore$\uparrow$}\\
%     \midrule
%     1 & 20.43$\pm$5.25 & 0.68$\pm$0.17 & 0.02$\pm$0.01 & 14.27
%       & \textbf{0.67} & 1.44
%       & 0.83 & 0.21 \\
%     3 & \textbf{20.95}$\pm$5.34 & \textbf{0.69}$\pm$0.16 & \textbf{0.02}$\pm$0.01 & \textbf{13.88}
%       & 0.69 & 1.44
%       & \textbf{0.81} & 0.20 \\
%     5 & 20.43$\pm$5.17 & 0.68$\pm$0.16 & 0.02$\pm$0.01 & 14.28
%       & 0.70 & \textbf{1.42}
%       & 0.83 & \textbf{0.21} \\
%     \bottomrule
%   \end{tabular}
%   \caption{Effect of Loss Weight}
%   \label{tab:loss_weight_ablation}
% \end{table*}

% Video + Audio table
\begin{table}[ht]
  \centering
  \footnotesize
  \setlength{\tabcolsep}{4pt}
  \renewcommand{\arraystretch}{0.95}
  \vspace{.5ex}
  \begin{tabular}{lcccccc}
    \toprule
    & \multicolumn{4}{c}{\textbf{Video Evaluation}} & \multicolumn{2}{c}{\textbf{Audio Evaluation}} \\
    \cmidrule(lr){2-5}\cmidrule(lr){6-7}
    \textbf{$\lambda$}
      & \textbf{PSNR$\uparrow$} & \textbf{SSIM$\uparrow$} & \textbf{LPIPS$\downarrow$} & \textbf{FVID$\downarrow$}
      & \textbf{FAD$\downarrow$} & \textbf{LSD$\downarrow$} \\
    \midrule
    1 & 20.43$\pm$5.25 & 0.68$\pm$0.17 & 0.02$\pm$0.01 & 14.27
      & \textbf{0.67} & 1.44 \\
    3 & \textbf{20.95}$\pm$5.34 & \textbf{0.69}$\pm$0.16 & \textbf{0.02}$\pm$0.01 & \textbf{13.88}
      & 0.69 & 1.44 \\
    5 & 20.43$\pm$5.17 & 0.68$\pm$0.16 & 0.02$\pm$0.01 & 14.28
      & 0.70 & \textbf{1.42} \\
    \bottomrule
  \end{tabular}
  \caption{Effect of $\lambda$ on removal performance}
  \label{tab:loss_weight_ablation_av}
\end{table}

% Synchronization-only table
\begin{table}[ht]
  \centering
  \footnotesize
  \setlength{\tabcolsep}{4pt}
  \renewcommand{\arraystretch}{0.95}
  \vspace{.5ex}
  \begin{tabular}{lcc}
    \toprule
    & \multicolumn{2}{c}{\textbf{Audiovisual Synchronization}} \\
    \cmidrule(lr){2-3}
    \textbf{$\lambda$}
      & \textbf{DeSyncScore$\downarrow$} & \textbf{ImageBindScore$\uparrow$} \\
    \midrule
    1 & 0.83 & 0.21 \\
    3 & \textbf{0.81} & 0.20 \\
    5 & 0.83 & \textbf{0.21} \\
    \bottomrule
  \end{tabular}
  \caption{Effect of $\lambda$ on audiovisual alignment}
  \label{tab:loss_weight_ablation_sync}
\end{table}

% \begin{table*}
%   \centering
%   \footnotesize
%   \setlength{\tabcolsep}{4pt}
%   \renewcommand{\arraystretch}{0.95}
%   \vspace{.5ex}
%   \resizebox{\textwidth}{!}{
%   \begin{tabular}{lrrcccccccc}
%     \toprule
%     & & & \multicolumn{4}{c}{\textbf{Video Evaluation}} & \multicolumn{2}{c}{\textbf{Audio Evaluation}} & \multicolumn{2}{c}{\textbf{Audiovisual Synchronization}}\\
%     \cmidrule(lr){4-7}\cmidrule(lr){8-9}\cmidrule(lr){10-11}
%     \textbf{Output Block} & \textbf{\# Head} & \textbf{\# Depth}
%       & \textbf{PSNR$\uparrow$} & \textbf{SSIM$\uparrow$} & \textbf{LPIPS$\downarrow$} & \textbf{FVID$\downarrow$}
%       & \textbf{FAD$\downarrow$} & \textbf{LSD$\downarrow$}
%       & \textbf{DeSyncScore$\downarrow$} & \textbf{ImageBindScore$\uparrow$}\\
%     \midrule
%     Linear Layer & 16 & 16
%       & 18.84$\pm$4.80 & 0.65$\pm$0.17 & 0.02$\pm$0.01 & 14.83
%       & 0.68 & \textbf{1.43}
%       & \textbf{0.82} & \textbf{0.21} \\
%     DiT Head & 8 & 8
%       & 20.03$\pm$5.16 & 0.67$\pm$0.17 & 0.02$\pm$0.03 & 14.66
%       & \textbf{0.65} & 1.44
%       & 0.84 & 0.21 \\
%     DiT Head & 16 & 16
%       & \textbf{20.43}$\pm$5.25 & \textbf{0.68}$\pm$0.17 & \textbf{0.02}$\pm$0.01 & \textbf{14.27}
%       & 0.67 & \textbf{1.43}
%       & 0.83 & 0.21 \\
%     \bottomrule
%   \end{tabular}%
%   }
%   \caption{\textbf{Effect of Disentangle Block.}}
%   \label{tab:dit_ablation}
% \end{table*}

\begin{table}
  \centering
  \scriptsize
  \setlength{\tabcolsep}{3pt}
  \renewcommand{\arraystretch}{0.9}
  \begin{tabular}{lccccccc}
    \toprule
    & \multicolumn{4}{c}{\textbf{Video Eval}} & \multicolumn{2}{c}{\textbf{Audio Eval}}\\
    \cmidrule(lr){2-5}\cmidrule(lr){6-7}
    \textbf{Block} & \textbf{PSNR$\uparrow$} & \textbf{SSIM$\uparrow$} & \textbf{LPIPS$\downarrow$} & \textbf{FVID$\downarrow$}
                   & \textbf{FAD$\downarrow$} & \textbf{LSD$\downarrow$} \\
    \midrule
    Linear (16H/16D)
      & 18.84$\pm$4.80 & 0.65$\pm$0.17 & 0.02$\pm$0.01 & 14.83
      & 0.68 & \textbf{1.43} \\
    DiT (8H/8D)
      & 20.03$\pm$5.16 & 0.67$\pm$0.17 & 0.02$\pm$0.03 & 14.66
      & \textbf{0.65} & 1.44 \\
    DiT (16H/16D)
      & \textbf{20.43}$\pm$5.25 & \textbf{0.68}$\pm$0.17 & \textbf{0.02}$\pm$0.01 & \textbf{14.27}
      & 0.67 & \textbf{1.43} \\
    \bottomrule
  \end{tabular}
  \caption{\textbf{Effect of audio head and video head on video and audio removal.} Here ``H/D'' in the block name denotes the number of attention heads and the number of DiT layers.}
  \label{tab:dit_ablation_va}
\end{table}

\paragraph{Effect of Video Head and Audio Head}\label{sec:disentangle}
In this experiment, we quantify the benefit of disentangling modality-specific heads with DiT blocks by comparing them to a single linear layer that directly truncates audio and video features. As shown in Table~\ref{tab:dit_ablation_sync}, while direct truncation yields slightly better temporal audiovisual synchronization and semantic correspondence. However, as shown in Table~\ref{tab:dit_ablation_va}, it results in lower PSNR/SSIM and higher LPIPS/FVID, indicating poorer output fidelity, 

\begin{table}
  \centering
  \footnotesize
  \setlength{\tabcolsep}{4pt}
  \renewcommand{\arraystretch}{0.95}
  \begin{tabular}{lrrcc}
    \toprule
    \textbf{Block} & \textbf{\# Heads} & \textbf{\# Depth}
      & \textbf{DeSyncScore$\downarrow$} & \textbf{ImageBindScore$\uparrow$} \\
    \midrule
    Linear Layer & 16 & 16
      & \textbf{0.82} & \textbf{0.21} \\
    DiT Head & 8 & 8
      & 0.84 & 0.21 \\
    DiT Head & 16 & 16
      & 0.83 & 0.21 \\
    \bottomrule
  \end{tabular}
  \caption{Effect of audio head and video head on audiovisual synchronization}
  \label{tab:dit_ablation_sync}
\end{table}

\subsection{Effect of CLAP Condition}
In this experiment, we study the effect of CLAP conditioning on audio editing. We compare two variants of our model in Table~\ref{tab:clap_condition_ablation}: one with CLAP conditioning and one without, using $\lambda = 1$ in both cases. With CLAP conditioning, PSNR and SSIM decrease, while FVID increases, indicating a reduction in visual fidelity. Meanwhile, FAD and LSD both decrease, indicating an improvement in audio quality. With CLAP conditioning, we also observe a slight decrease in DeSyncScore, suggesting better audiovisual temporal synchronization. Overall, these results show that CLAP conditioning improves audio editing and synchronization in our joint editing model, at the cost of some visual quality.

% \paul{check if these are the only ablations... the pipeline overall looks complicated there must be other components that need ablating}

% \subsection{Effect of Condition}
% In this experiment, we aim to test the effect of audio clap condition. 
% \input{sec/8_limitations}
\section{Conclusion}
In this work, we present SAVEBench, a new dataset generated by our construction pipeline to support the audiovisual joint removal task. Using SAVEBench, we train the Schr\"{o}dinger Audio-Visual Editor (SAVE), a joint audiovisual generative model powered by the  \sbfm/. Our experiments show that SAVE achieves performance comparable to single-modality editors, while delivering stronger temporal synchronization, better audiovisual semantic alignment, and higher perceptual quality.
\paragraph{Limitations and Broader Impact}
In this work, we use two objects for dataset construction and employ our synthetic pipeline to produce paired samples for object-level joint audiovisual removal. However, our model architecture can be applied to any paired audiovisual dataset. To demonstrate its generalizability, we evaluate on real-world audiovisual data and on videos containing three or more objects, removing multiple objects in our supplementary experiments. Furthermore, our paired dataset naturally supports object addition and can be extended to finer-grained audiovisual stylization. We hope this work paves the way for a new paradigm in joint audiovisual stylization.
\clearpage
{
    \small
    \bibliographystyle{ieeenat_fullname}
    \bibliography{main}

@String(CVPR= {IEEE Conf. Comput. Vis. Pattern Recog.})

@String(ICCV= {Int. Conf. Comput. Vis.})

@String(NIPS= {Adv. Neural Inform. Process. Syst.})

@String(ICASSP=	{ICASSP})

@String(ICLR = {Int. Conf. Learn. Represent.})

@String(AAAI = {AAAI})

@String(CVPR  = {CVPR})

@String(ICCV  = {ICCV})

@String(NIPS  = {NeurIPS})

@String(ICLR  = {ICLR})

@misc{li2022endtoendframeworkflowguidedvideo,
      title={Towards An End-to-End Framework for Flow-Guided Video Inpainting}, 
      author={Zhen Li and Cheng-Ze Lu and Jianhua Qin and Chun-Le Guo and Ming-Ming Cheng},
      year={2022},
      eprint={2204.02663},
      archivePrefix={arXiv},
      primaryClass={eess.IV},
      url={https://arxiv.org/abs/2204.02663}, 
}

@misc{zhou2023propainterimprovingpropagationtransformer,
      title={ProPainter: Improving Propagation and Transformer for Video Inpainting}, 
      author={Shangchen Zhou and Chongyi Li and Kelvin C. K. Chan and Chen Change Loy},
      year={2023},
      eprint={2309.03897},
      archivePrefix={arXiv},
      primaryClass={cs.CV},
      url={https://arxiv.org/abs/2309.03897}, 
}

@misc{liu2024audioldm2learningholistic,
      title={AudioLDM 2: Learning Holistic Audio Generation with Self-supervised Pretraining}, 
      author={Haohe Liu and Yi Yuan and Xubo Liu and Xinhao Mei and Qiuqiang Kong and Qiao Tian and Yuping Wang and Wenwu Wang and Yuxuan Wang and Mark D. Plumbley},
      year={2024},
      eprint={2308.05734},
      archivePrefix={arXiv},
      primaryClass={cs.SD},
      url={https://arxiv.org/abs/2308.05734}, 
}

@article{liang2024foundations,
  title={Foundations \& trends in multimodal machine learning: Principles, challenges, and open questions},
  author={Liang, Paul Pu and Zadeh, Amir and Morency, Louis-Philippe},
  journal={ACM Computing Surveys},
  volume={56},
  number={10},
  pages={1--42},
  year={2024},
  publisher={ACM New York, NY}
}

@misc{manor2024zeroshotunsupervisedtextbasedaudio,
      title={Zero-Shot Unsupervised and Text-Based Audio Editing Using DDPM Inversion}, 
      author={Hila Manor and Tomer Michaeli},
      year={2024},
      eprint={2402.10009},
      archivePrefix={arXiv},
      primaryClass={cs.SD},
      url={https://arxiv.org/abs/2402.10009}, 
}

@misc{wasserman2025paintinpaintlearningadd,
      title={Paint by Inpaint: Learning to Add Image Objects by Removing Them First}, 
      author={Navve Wasserman and Noam Rotstein and Roy Ganz and Ron Kimmel},
      year={2025},
      eprint={2404.18212},
      archivePrefix={arXiv},
      primaryClass={cs.CV},
      url={https://arxiv.org/abs/2404.18212}, 
}

@book{oppenheim1999dtsp,
  title     = {Discrete-Time Signal Processing},
  author    = {Oppenheim, Alan V. and Schafer, Ronald W. and Buck, John R.},
  edition   = {2},
  publisher = {Prentice Hall},
  year      = {1999}
}

@misc{wisdom2020whatsfussfreeuniversal,
      title={What's All the FUSS About Free Universal Sound Separation Data?}, 
      author={Scott Wisdom and Hakan Erdogan and Daniel Ellis and Romain Serizel and Nicolas Turpault and Eduardo Fonseca and Justin Salamon and Prem Seetharaman and John Hershey},
      year={2020},
      eprint={2011.00803},
      archivePrefix={arXiv},
      primaryClass={cs.SD},
      url={https://arxiv.org/abs/2011.00803}, 
}

@misc{henschel2025streamingt2vconsistentdynamicextendable,
      title={StreamingT2V: Consistent, Dynamic, and Extendable Long Video Generation from Text}, 
      author={Roberto Henschel and Levon Khachatryan and Hayk Poghosyan and Daniil Hayrapetyan and Vahram Tadevosyan and Zhangyang Wang and Shant Navasardyan and Humphrey Shi},
      year={2025},
      eprint={2403.14773},
      archivePrefix={arXiv},
      primaryClass={cs.CV},
      url={https://arxiv.org/abs/2403.14773}, 
}

@misc{fu2025objectaveditobjectlevelaudiovisualediting,
      title={Object-AVEdit: An Object-level Audio-Visual Editing Model}, 
      author={Youquan Fu and Ruiyang Si and Hongfa Wang and Dongzhan Zhou and Jiacheng Sun and Ping Luo and Di Hu and Hongyuan Zhang and Xuelong Li},
      year={2025},
      eprint={2510.00050},
      archivePrefix={arXiv},
      primaryClass={cs.MM},
      url={https://arxiv.org/abs/2510.00050}, 
}

@misc{girdhar2023imagebindembeddingspacebind,
      title={ImageBind: One Embedding Space To Bind Them All}, 
      author={Rohit Girdhar and Alaaeldin El-Nouby and Zhuang Liu and Mannat Singh and Kalyan Vasudev Alwala and Armand Joulin and Ishan Misra},
      year={2023},
      eprint={2305.05665},
      archivePrefix={arXiv},
      primaryClass={cs.CV},
      url={https://arxiv.org/abs/2305.05665}, 
}

@misc{iashin2024synchformerefficientsynchronizationsparse,
      title={Synchformer: Efficient Synchronization from Sparse Cues}, 
      author={Vladimir Iashin and Weidi Xie and Esa Rahtu and Andrew Zisserman},
      year={2024},
      eprint={2401.16423},
      archivePrefix={arXiv},
      primaryClass={cs.CV},
      url={https://arxiv.org/abs/2401.16423}, 
}

@article{liu2023grounding,
  title={Grounding dino: Marrying dino with grounded pre-training for open-set object detection},
  author={Liu, Shilong and Zeng, Zhaoyang and Ren, Tianhe and Li, Feng and Zhang, Hao and Yang, Jie and Li, Chunyuan and Yang, Jianwei and Su, Hang and Zhu, Jun and others},
  journal={arXiv preprint arXiv:2303.05499},
  year={2023}
}

@misc{wu2024largescalecontrastivelanguageaudiopretraining,
      title={Large-scale Contrastive Language-Audio Pretraining with Feature Fusion and Keyword-to-Caption Augmentation}, 
      author={Yusong Wu and Ke Chen and Tianyu Zhang and Yuchen Hui and Marianna Nezhurina and Taylor Berg-Kirkpatrick and Shlomo Dubnov},
      year={2024},
      eprint={2211.06687},
      archivePrefix={arXiv},
      primaryClass={cs.SD},
      url={https://arxiv.org/abs/2211.06687}, 
}

@misc{wang2023auditaudioeditingfollowing,
      title={AUDIT: Audio Editing by Following Instructions with Latent Diffusion Models}, 
      author={Yuancheng Wang and Zeqian Ju and Xu Tan and Lei He and Zhizheng Wu and Jiang Bian and Sheng Zhao},
      year={2023},
      eprint={2304.00830},
      archivePrefix={arXiv},
      primaryClass={cs.SD},
      url={https://arxiv.org/abs/2304.00830}, 
}

@misc{Valentini2017VoiceBankDEMAND,
  author       = {Valentini-Botinhao, Cassia},
  title        = {Noisy Speech Database for Training Speech Enhancement Algorithms and TTS Models},
  howpublished = {University of Edinburgh DataShare},
  year         = {2017},
  doi          = {10.7488/ds/2117},
  url          = {https://datashare.ed.ac.uk/handle/10283/2791}
}

@misc{hong2023lvosbenchmarklongtermvideo,
      title={LVOS: A Benchmark for Long-term Video Object Segmentation}, 
      author={Lingyi Hong and Wenchao Chen and Zhongying Liu and Wei Zhang and Pinxue Guo and Zhaoyu Chen and Wenqiang Zhang},
      year={2023},
      eprint={2211.10181},
      archivePrefix={arXiv},
      primaryClass={cs.CV},
      url={https://arxiv.org/abs/2211.10181}, 
}

@misc{cosentino2020librimixopensourcedatasetgeneralizable,
      title={LibriMix: An Open-Source Dataset for Generalizable Speech Separation}, 
      author={Joris Cosentino and Manuel Pariente and Samuele Cornell and Antoine Deleforge and Emmanuel Vincent},
      year={2020},
      eprint={2005.11262},
      archivePrefix={arXiv},
      primaryClass={eess.AS},
      url={https://arxiv.org/abs/2005.11262}, 
}

@misc{ding2023mosenewdatasetvideo,
      title={MOSE: A New Dataset for Video Object Segmentation in Complex Scenes}, 
      author={Henghui Ding and Chang Liu and Shuting He and Xudong Jiang and Philip H. S. Torr and Song Bai},
      year={2023},
      eprint={2302.01872},
      archivePrefix={arXiv},
      primaryClass={cs.CV},
      url={https://arxiv.org/abs/2302.01872}, 
}

@inproceedings{Perazzi2016DAVIS,
  author    = {Perazzi, Federico and Pont-Tuset, Jordi and McWilliams, Brian and Van Gool, Luc and Gross, Markus and Sorkine-Hornung, Alexander},
  title     = {A Benchmark Dataset and Evaluation Methodology for Video Object Segmentation},
  booktitle = {CVPR},
  year      = {2016},
  url       = {https://openaccess.thecvf.com/content_cvpr_2016/html/Perazzi_A_Benchmark_Dataset_CVPR_2016_paper.html}
}

@misc{lee2021acav100mautomaticcurationlargescale,
      title={ACAV100M: Automatic Curation of Large-Scale Datasets for Audio-Visual Video Representation Learning}, 
      author={Sangho Lee and Jiwan Chung and Youngjae Yu and Gunhee Kim and Thomas Breuel and Gal Chechik and Yale Song},
      year={2021},
      eprint={2101.10803},
      archivePrefix={arXiv},
      primaryClass={cs.CV},
      url={https://arxiv.org/abs/2101.10803}, 
}

@article{audiovisual_survey,
author = {Vila\c{c}a, Lu\'{\i}s and Yu, Yi and Viana, Paula},
title = {A Survey of Recent Advances and Challenges in Deep Audio-Visual Correlation Learning},
year = {2025},
issue_date = {December 2025},
publisher = {Association for Computing Machinery},
address = {New York, NY, USA},
volume = {57},
number = {12},
issn = {0360-0300},
url = {https://doi.org/10.1145/3696445},
doi = {10.1145/3696445},
journal = {ACM Comput. Surv.},
month = jul,
articleno = {299},
numpages = {46}
}

@article{correia2021avui,
  title   = {From GUI to AVUI: Situating Audiovisual User Interfaces Within Human-Computer Interaction and Related Fields},
  author  = {Correia, Nuno N. and Tanaka, Atau},
  journal = {EAI Endorsed Transactions on Creative Technologies},
  volume  = {8},
  number  = {27},
  pages   = {e5},
  year    = {2021},
  doi     = {10.4108/eai.12-5-2021.169913}
}

@misc{brooks2023instructpix2pixlearningfollowimage,
      title={InstructPix2Pix: Learning to Follow Image Editing Instructions}, 
      author={Tim Brooks and Aleksander Holynski and Alexei A. Efros},
      year={2023},
      eprint={2211.09800},
      archivePrefix={arXiv},
      primaryClass={cs.CV},
      url={https://arxiv.org/abs/2211.09800}, 
}

@inproceedings{cheng2024taming,
  title={MMAudio: Taming Multimodal Joint Training for High-Quality Video-to-Audio Synthesis},
  author={Cheng, Ho Kei and Ishii, Masato and Hayakawa, Akio and Shibuya, Takashi and Schwing, Alexander and Mitsufuji, Yuki},
  booktitle={CVPR},
  year={2025}
}

@inproceedings{saito2025soundctm,
  title={Sound{CTM}: Unifying Score-based and Consistency Models for Full-band Text-to-Sound Generation},
  author={Koichi Saito and Dongjun Kim and Takashi Shibuya and Chieh-Hsin Lai and Zhi Zhong and Yuhta Takida and Yuki Mitsufuji},
  booktitle={The Thirteenth International Conference on Learning Representations},
  year={2025},
  url={https://openreview.net/forum?id=KrK6zXbjfO}
}

@misc{wu2024languagedrivenvideoinpaintingmultimodal,
      title={Towards Language-Driven Video Inpainting via Multimodal Large Language Models}, 
      author={Jianzong Wu and Xiangtai Li and Chenyang Si and Shangchen Zhou and Jingkang Yang and Jiangning Zhang and Yining Li and Kai Chen and Yunhai Tong and Ziwei Liu and Chen Change Loy},
      year={2024},
      eprint={2401.10226},
      archivePrefix={arXiv},
      primaryClass={cs.CV},
      url={https://arxiv.org/abs/2401.10226}, 
}

@article{zhao2024cvvae,
  title={CV-VAE: A Compatible Video VAE for Latent Generative Video Models},
  author={Zhao, Sijie and Zhang, Yong and Cun, Xiaodong and Yang, Shaoshu and Niu, Muyao and Li, Xiaoyu and Hu, Wenbo and Shan, Ying},
  journal={https://arxiv.org/abs/2405.20279},
  year={2024}
}

@article{yu2023inpaint,
  title={Inpaint Anything: Segment Anything Meets Image Inpainting},
  author={Yu, Tao and Feng, Runseng and Feng, Ruoyu and Liu, Jinming and Jin, Xin and Zeng, Wenjun and Chen, Zhibo},
  journal={arXiv preprint arXiv:2304.06790},
  year={2023}
}

@misc{bian2025videopainteranylengthvideoinpainting,
      title={VideoPainter: Any-length Video Inpainting and Editing with Plug-and-Play Context Control}, 
      author={Yuxuan Bian and Zhaoyang Zhang and Xuan Ju and Mingdeng Cao and Liangbin Xie and Ying Shan and Qiang Xu},
      year={2025},
      eprint={2503.05639},
      archivePrefix={arXiv},
      primaryClass={cs.CV},
      url={https://arxiv.org/abs/2503.05639}, 
}

@misc{ravi2024sam2segmentimages,
      title={SAM 2: Segment Anything in Images and Videos}, 
      author={Nikhila Ravi and Valentin Gabeur and Yuan-Ting Hu and Ronghang Hu and Chaitanya Ryali and Tengyu Ma and Haitham Khedr and Roman Rädle and Chloe Rolland and Laura Gustafson and Eric Mintun and Junting Pan and Kalyan Vasudev Alwala and Nicolas Carion and Chao-Yuan Wu and Ross Girshick and Piotr Dollár and Christoph Feichtenhofer},
      year={2024},
      eprint={2408.00714},
      archivePrefix={arXiv},
      primaryClass={cs.CV},
      url={https://arxiv.org/abs/2408.00714}, 
}

@inproceedings{Wang2018Vid2Vid,
  author    = {Ting{-}Chun Wang and Ming{-}Yu Liu and Jun{-}Yan Zhu and Guilin Liu and Andrew Tao and Jan Kautz and Bryan Catanzaro},
  title     = {Video-to-Video Synthesis},
  booktitle = {Advances in Neural Information Processing Systems},
  volume    = {31},
  year      = {2018},
  publisher = {Curran Associates, Inc.},
  url       = {https://papers.nips.cc/paper_files/paper/2018/hash/d86ea612dec96096c5e0fcc8dd42ab6d-Abstract.html},
  eprint    = {1808.06601},
  archivePrefix = {arXiv}
}

@article{Qwen-Audio,
  title={Qwen-Audio: Advancing Universal Audio Understanding via Unified Large-Scale Audio-Language Models},
  author={Chu, Yunfei and Xu, Jin and Zhou, Xiaohuan and Yang, Qian and Zhang, Shiliang and Yan, Zhijie  and Zhou, Chang and Zhou, Jingren},
  journal={arXiv preprint arXiv:2311.07919},
  year={2023}
}

@inproceedings{zhou2023propainter,
   title={{ProPainter}: Improving Propagation and Transformer for Video Inpainting},
   author={Zhou, Shangchen and Li, Chongyi and Chan, Kelvin C.K and Loy, Chen Change},
   booktitle={Proceedings of IEEE International Conference on Computer Vision (ICCV)},
   year={2023}
}

@misc{jiang2025vaceallinonevideocreation,
      title={VACE: All-in-One Video Creation and Editing}, 
      author={Zeyinzi Jiang and Zhen Han and Chaojie Mao and Jingfeng Zhang and Yulin Pan and Yu Liu},
      year={2025},
      eprint={2503.07598},
      archivePrefix={arXiv},
      primaryClass={cs.CV},
      url={https://arxiv.org/abs/2503.07598}, 
}

@misc{lin2025zeroshotaudiovisualeditingcrossmodal,
      title={Zero-Shot Audio-Visual Editing via Cross-Modal Delta Denoising}, 
      author={Yan-Bo Lin and Kevin Lin and Zhengyuan Yang and Linjie Li and Jianfeng Wang and Chung-Ching Lin and Xiaofei Wang and Gedas Bertasius and Lijuan Wang},
      year={2025},
      eprint={2503.20782},
      archivePrefix={arXiv},
      primaryClass={cs.CV},
      url={https://arxiv.org/abs/2503.20782}, 
}

@misc{liang2024languageguidedjointaudiovisualediting,
      title={Language-Guided Joint Audio-Visual Editing via One-Shot Adaptation}, 
      author={Susan Liang and Chao Huang and Yapeng Tian and Anurag Kumar and Chenliang Xu},
      year={2024},
      eprint={2410.07463},
      archivePrefix={arXiv},
      primaryClass={cs.CV},
      url={https://arxiv.org/abs/2410.07463}, 
}

@article{Qwen-VL,
  title={Qwen-VL: A Versatile Vision-Language Model for Understanding, Localization, Text Reading, and Beyond},
  author={Bai, Jinze and Bai, Shuai and Yang, Shusheng and Wang, Shijie and Tan, Sinan and Wang, Peng and Lin, Junyang and Zhou, Chang and Zhou, Jingren},
  journal={arXiv preprint arXiv:2308.12966},
  year={2023}
}

@inproceedings{cheng2025taming,
  title={{MMAudio}: Taming Multimodal Joint Training for High-Quality Video-to-Audio Synthesis},
  author={Cheng, Ho Kei and Ishii, Masato and Hayakawa, Akio and Shibuya, Takashi and Schwing, Alexander and Mitsufuji, Yuki},
  booktitle={CVPR},
  year={2025}
}

@misc{wang2025klingfoleymultimodaldiffusiontransformer,
      title={Kling-Foley: Multimodal Diffusion Transformer for High-Quality Video-to-Audio Generation}, 
      author={Jun Wang and Xijuan Zeng and Chunyu Qiang and Ruilong Chen and Shiyao Wang and Le Wang and Wangjing Zhou and Pengfei Cai and Jiahui Zhao and Nan Li and Zihan Li and Yuzhe Liang and Xiaopeng Wang and Haorui Zheng and Ming Wen and Kang Yin and Yiran Wang and Nan Li and Feng Deng and Liang Dong and Chen Zhang and Di Zhang and Kun Gai},
      year={2025},
      eprint={2506.19774},
      archivePrefix={arXiv},
      primaryClass={eess.AS},
      url={https://arxiv.org/abs/2506.19774}, 
}

@inproceedings{gemmeke2017audioset,
  title        = {Audio Set: An ontology and human-labeled dataset for audio events},
  author       = {Gemmeke, Jort F. and Ellis, Daniel P. W. and Freedman, Dylan and Jansen, Aren and Lawrence, Wade and Moore, R. Channing and Plakal, Manoj and Ritter, Marvin},
  booktitle    = {2017 IEEE International Conference on Acoustics, Speech and Signal Processing (ICASSP)},
  pages        = {776--780},
  year         = {2017},
  address      = {New Orleans, LA, USA},
  publisher    = {IEEE},
  doi          = {10.1109/ICASSP.2017.7952261}
}

@article{wang2023audit,
  title={Audit: Audio editing by following instructions with latent diffusion models},
  author={Wang, Yuancheng and Ju, Zeqian and Tan, Xu and He, Lei and Wu, Zhizheng and Bian, Jiang and others},
  journal={Advances in Neural Information Processing Systems},
  volume={36},
  pages={71340--71357},
  year={2023}
}

@inproceedings{cheng2025audio,
  title={Audio Texture Manipulation by Exemplar-Based Analogy},
  author={Cheng, Kan Jen and Li, Tingle and Anumanchipalli, Gopala},
  booktitle={ICASSP 2025-2025 IEEE International Conference on Acoustics, Speech and Signal Processing (ICASSP)},
  pages={1--5},
  year={2025},
  organization={IEEE}
}

@article{liang2025audiomorphix,
  title={AudioMorphix: Training-free audio editing with diffusion probabilistic models},
  author={Liang, Jinhua and Chen, Yuanzhe and Yuan, Yi and Jia, Dongya and Zhuang, Xiaobin and Chen, Zhuo and Wang, Yuping and Wang, Yuxuan},
  journal={arXiv preprint arXiv:2505.16076},
  year={2025}
}

@inproceedings{sung2023sound,
  title={Sound to visual scene generation by audio-to-visual latent alignment},
  author={Sung-Bin, Kim and Senocak, Arda and Ha, Hyunwoo and Owens, Andrew and Oh, Tae-Hyun},
  booktitle={Proceedings of the IEEE/CVF Conference on Computer Vision and Pattern Recognition},
  pages={6430--6440},
  year={2023}
}

@inproceedings{li2022learning,
  title={Learning visual styles from audio-visual associations},
  author={Li, Tingle and Liu, Yichen and Owens, Andrew and Zhao, Hang},
  booktitle={European Conference on Computer Vision},
  pages={235--252},
  year={2022},
  organization={Springer}
}

@article{iashin2021taming,
  title={Taming visually guided sound generation},
  author={Iashin, Vladimir and Rahtu, Esa},
  journal={arXiv preprint arXiv:2110.08791},
  year={2021}
}

@inproceedings{pascual2024masked,
  title={Masked generative video-to-audio transformers with enhanced synchronicity},
  author={Pascual, Santiago and Yeh, Chunghsin and Tsiamas, Ioannis and Serr{\`a}, Joan},
  booktitle={European Conference on Computer Vision},
  pages={247--264},
  year={2024},
  organization={Springer}
}

@inproceedings{su2023physics,
  title={Physics-driven diffusion models for impact sound synthesis from videos},
  author={Su, Kun and Qian, Kaizhi and Shlizerman, Eli and Torralba, Antonio and Gan, Chuang},
  booktitle={Proceedings of the IEEE/CVF Conference on Computer Vision and Pattern Recognition},
  pages={9749--9759},
  year={2023}
}

@inproceedings{wang2024v2a,
  title={V2a-mapper: A lightweight solution for vision-to-audio generation by connecting foundation models},
  author={Wang, Heng and Ma, Jianbo and Pascual, Santiago and Cartwright, Richard and Cai, Weidong},
  booktitle={Proceedings of the AAAI Conference on Artificial Intelligence},
  volume={38},
  number={14},
  pages={15492--15501},
  year={2024}
}

@article{yang2024draw,
  title={Draw an audio: Leveraging multi-instruction for video-to-audio synthesis},
  author={Yang, Qi and Mao, Binjie and Wang, Zili and Nie, Xing and Gao, Pengfei and Guo, Ying and Zhen, Cheng and Yan, Pengfei and Xiang, Shiming},
  journal={arXiv preprint arXiv:2409.06135},
  year={2024}
}

@inproceedings{comunita2024syncfusion,
  title={Syncfusion: Multimodal onset-synchronized video-to-audio foley synthesis},
  author={Comunit{\`a}, Marco and Gramaccioni, Riccardo F and Postolache, Emilian and Rodol{\`a}, Emanuele and Comminiello, Danilo and Reiss, Joshua D},
  booktitle={ICASSP 2024-2024 IEEE International Conference on Acoustics, Speech and Signal Processing (ICASSP)},
  pages={936--940},
  year={2024},
  organization={IEEE}
}

@article{hu2024video,
  title={Video-to-audio generation with fine-grained temporal semantics},
  author={Hu, Yuchen and Gu, Yu and Li, Chenxing and Chen, Rilin and Yu, Dong},
  journal={arXiv preprint arXiv:2409.14709},
  year={2024}
}

@inproceedings{jeong2025read,
  title={Read, watch and scream! sound generation from text and video},
  author={Jeong, Yujin and Kim, Yunji and Chun, Sanghyuk and Lee, Jiyoung},
  booktitle={Proceedings of the AAAI Conference on Artificial Intelligence},
  volume={39},
  number={17},
  pages={17590--17598},
  year={2025}
}

@inproceedings{jeong2023power,
  title={The power of sound (tpos): Audio reactive video generation with stable diffusion},
  author={Jeong, Yujin and Ryoo, Wonjeong and Lee, Seunghyun and Seo, Dabin and Byeon, Wonmin and Kim, Sangpil and Kim, Jinkyu},
  booktitle={Proceedings of the IEEE/CVF International Conference on Computer Vision},
  pages={7822--7832},
  year={2023}
}

@inproceedings{yariv2024diverse,
  title={Diverse and aligned audio-to-video generation via text-to-video model adaptation},
  author={Yariv, Guy and Gat, Itai and Benaim, Sagie and Wolf, Lior and Schwartz, Idan and Adi, Yossi},
  booktitle={Proceedings of the AAAI Conference on Artificial Intelligence},
  volume={38},
  number={7},
  pages={6639--6647},
  year={2024}
}

@inproceedings{zhang2024audio,
  title={Audio-synchronized visual animation},
  author={Zhang, Lin and Mo, Shentong and Zhang, Yijing and Morgado, Pedro},
  booktitle={European Conference on Computer Vision},
  pages={1--18},
  year={2024},
  organization={Springer}
}

@article{zhao2025uniform,
  title={UniForm: A Unified Multi-Task Diffusion Transformer for Audio-Video Generation},
  author={Zhao, Lei and Feng, Linfeng and Ge, Dongxu and Chen, Rujin and Yi, Fangqiu and Zhang, Chi and Zhang, Xiao-Lei and Li, Xuelong},
  journal={arXiv preprint arXiv:2502.03897},
  year={2025}
}

@article{liu2024syncflow,
  title={SyncFlow: Toward Temporally Aligned Joint Audio-Video Generation from Text},
  author={Liu, Haohe and Lan, Gael Le and Mei, Xinhao and Ni, Zhaoheng and Kumar, Anurag and Nagaraja, Varun and Wang, Wenwu and Plumbley, Mark D and Shi, Yangyang and Chandra, Vikas},
  journal={arXiv preprint arXiv:2412.15220},
  year={2024}
}

@article{tang2023any,
  title={Any-to-any generation via composable diffusion},
  author={Tang, Zineng and Yang, Ziyi and Zhu, Chenguang and Zeng, Michael and Bansal, Mohit},
  journal={Advances in Neural Information Processing Systems},
  volume={36},
  pages={16083--16099},
  year={2023}
}

@inproceedings{liang2024language,
  title={Language-guided joint audio-visual editing via one-shot adaptation},
  author={Liang, Susan and Huang, Chao and Tian, Yapeng and Kumar, Anurag and Xu, Chenliang},
  booktitle={Proceedings of the Asian Conference on Computer Vision},
  pages={1011--1027},
  year={2024}
}

@misc{zhang2018unreasonableeffectivenessdeepfeatures,
      title={The Unreasonable Effectiveness of Deep Features as a Perceptual Metric}, 
      author={Richard Zhang and Phillip Isola and Alexei A. Efros and Eli Shechtman and Oliver Wang},
      year={2018},
      eprint={1801.03924},
      archivePrefix={arXiv},
      primaryClass={cs.CV},
      url={https://arxiv.org/abs/1801.03924}, 
}

@ARTICLE{SSIM,
  author={Zhou Wang and Bovik, A.C. and Sheikh, H.R. and Simoncelli, E.P.},
  journal={IEEE Transactions on Image Processing}, 
  title={Image quality assessment: from error visibility to structural similarity}, 
  year={2004},
  volume={13},
  number={4},
  pages={600-612},
  doi={10.1109/TIP.2003.819861}}

@article{lin2025zero,
  title={Zero-Shot Audio-Visual Editing via Cross-Modal Delta Denoising},
  author={Lin, Yan-Bo and Lin, Kevin and Yang, Zhengyuan and Li, Linjie and Wang, Jianfeng and Lin, Chung-Ching and Wang, Xiaofei and Bertasius, Gedas and Wang, Lijuan},
  journal={arXiv preprint arXiv:2503.20782},
  year={2025}
}

@INPROCEEDINGS{PSNRSSIM,
  author={Horé, Alain and Ziou, Djemel},
  booktitle={2010 20th International Conference on Pattern Recognition}, 
  title={Image Quality Metrics: PSNR vs. SSIM}, 
  year={2010},
  volume={},
  number={},
  pages={2366-2369},
  doi={10.1109/ICPR.2010.579}}

@article{Chen2014OnTR,
  title={On the Relation Between Optimal Transport and Schr{\"o}dinger Bridges: A Stochastic Control Viewpoint},
  author={Yongxin Chen and Tryphon T. Georgiou and Michele Pavon},
  journal={Journal of Optimization Theory and Applications},
  year={2014},
  volume={169},
  pages={671-691},
  url={https://api.semanticscholar.org/CorpusID:8968928}
}

@misc{heinrich2025radiov25improvedbaselinesagglomerative,
      title={RADIOv2.5: Improved Baselines for Agglomerative Vision Foundation Models}, 
      author={Greg Heinrich and Mike Ranzinger and Hongxu and Yin and Yao Lu and Jan Kautz and Andrew Tao and Bryan Catanzaro and Pavlo Molchanov},
      year={2025},
      eprint={2412.07679},
      archivePrefix={arXiv},
      primaryClass={cs.CV},
      url={https://arxiv.org/abs/2412.07679}, 
}

@inproceedings{ho2020denoising,
  title={Denoising diffusion probabilistic models},
  author={Jonathan Ho and Ajay Jain and Pieter Abbeel},
  booktitle=NIPS,
  volume={33},
  pages={6840--6851},
  year={2020}
}

@inproceedings{lipman2023flow,
title={Flow Matching for Generative Modeling},
author={Yaron Lipman and Ricky T. Q. Chen and Heli Ben-Hamu and Maximilian Nickel and Matthew Le},
booktitle=ICLR,
year={2023},
url={https://openreview.net/forum?id=PqvMRDCJT9t}
}

@article{Schrodinger1932,
  author    = {Erwin Schr{\"o}dinger},
  title     = {Sur la théorie relativiste de l’électron et l’interprétation de la mécanique quantique},
  journal   = {Annales de l'Institut Henri Poincaré},
  volume    = {2},
  number    = {4},
  year      = {1932},
  pages     = {269--310}
}

@inproceedings{sohl-dickstein2015deep,
  title={Deep unsupervised learning using nonequilibrium thermodynamics},
  author={Jascha Sohl-Dickstein and Eric Weiss and Niru Maheswaranathan and Surya Ganguli},
  booktitle={Proceedings of the 32nd International Conference on Machine Learning},
  editor={Francis Bach and David Blei},
  volume={37},
  series={Proceedings of Machine Learning Research},
  pages={2256--2265},
  year={2015},
  address={Lille, France},
  month={7},
  publisher={PMLR}
}

@inproceedings{Song2019generative,
  title={Generative Modeling by Estimating Gradients of the Data Distribution},
  author={Yang Song and Stefano Ermon},
  booktitle=NIPS,
  year={2019},
  url={https://arxiv.org/pdf/1907.05600},
}

@inproceedings{Song2019sliced,
  title={Sliced Score Matching: A Scalable Approach to Density and Score Estimation},
  author={Yang Song and Sahaj Garg and Jiaxin Shi and Stefano Ermon},
  booktitle={Uncertainty in Artificial Intelligence (UAI)},
  year={2019},
}

@inproceedings{Song2020improved,
  title={Improved Techniques for Training Score-Based Generative Models},
  author={Yang Song and Stefano Ermon},
  booktitle=NIPS,
  year={2020},
  url={ https://arxiv.org/pdf/2006.09011},
}

@inproceedings{Song2021score,
  title={Score-Based Generative Modeling through Stochastic Differential Equations},
  author={Yang Song and Jascha Sohl-Dickstein and Diederik P. Kingma and Abhishek Kumar and Stefano Ermon and Ben Poole},
  booktitle=ICLR,
  year={2021},
  url={https://openreview.net/forum?id=PxTIG12RRHS}
}

@misc{openai2024gpt4ocard,
      title={GPT-4o System Card}, 
      author={OpenAI and : and Aaron Hurst and Adam Lerer and Adam P. Goucher and Adam Perelman and Aditya Ramesh and Aidan Clark and AJ Ostrow and Akila Welihinda and Alan Hayes and Alec Radford and Aleksander Mądry and Alex Baker-Whitcomb and Alex Beutel and Alex Borzunov and Alex Carney and Alex Chow and Alex Kirillov and Alex Nichol and Alex Paino and Alex Renzin and Alex Tachard Passos and Alexander Kirillov and Alexi Christakis and Alexis Conneau and Ali Kamali and Allan Jabri and Allison Moyer and Allison Tam and Amadou Crookes and Amin Tootoochian and Amin Tootoonchian and Ananya Kumar and Andrea Vallone and Andrej Karpathy and Andrew Braunstein and Andrew Cann and Andrew Codispoti and Andrew Galu and Andrew Kondrich and Andrew Tulloch and Andrey Mishchenko and Angela Baek and Angela Jiang and Antoine Pelisse and Antonia Woodford and Anuj Gosalia and Arka Dhar and Ashley Pantuliano and Avi Nayak and Avital Oliver and Barret Zoph and Behrooz Ghorbani and Ben Leimberger and Ben Rossen and Ben Sokolowsky and Ben Wang and Benjamin Zweig and Beth Hoover and Blake Samic and Bob McGrew and Bobby Spero and Bogo Giertler and Bowen Cheng and Brad Lightcap and Brandon Walkin and Brendan Quinn and Brian Guarraci and Brian Hsu and Bright Kellogg and Brydon Eastman and Camillo Lugaresi and Carroll Wainwright and Cary Bassin and Cary Hudson and Casey Chu and Chad Nelson and Chak Li and Chan Jun Shern and Channing Conger and Charlotte Barette and Chelsea Voss and Chen Ding and Cheng Lu and Chong Zhang and Chris Beaumont and Chris Hallacy and Chris Koch and Christian Gibson and Christina Kim and Christine Choi and Christine McLeavey and Christopher Hesse and Claudia Fischer and Clemens Winter and Coley Czarnecki and Colin Jarvis and Colin Wei and Constantin Koumouzelis and Dane Sherburn and Daniel Kappler and Daniel Levin and Daniel Levy and David Carr and David Farhi and David Mely and David Robinson and David Sasaki and Denny Jin and Dev Valladares and Dimitris Tsipras and Doug Li and Duc Phong Nguyen and Duncan Findlay and Edede Oiwoh and Edmund Wong and Ehsan Asdar and Elizabeth Proehl and Elizabeth Yang and Eric Antonow and Eric Kramer and Eric Peterson and Eric Sigler and Eric Wallace and Eugene Brevdo and Evan Mays and Farzad Khorasani and Felipe Petroski Such and Filippo Raso and Francis Zhang and Fred von Lohmann and Freddie Sulit and Gabriel Goh and Gene Oden and Geoff Salmon and Giulio Starace and Greg Brockman and Hadi Salman and Haiming Bao and Haitang Hu and Hannah Wong and Haoyu Wang and Heather Schmidt and Heather Whitney and Heewoo Jun and Hendrik Kirchner and Henrique Ponde de Oliveira Pinto and Hongyu Ren and Huiwen Chang and Hyung Won Chung and Ian Kivlichan and Ian O'Connell and Ian O'Connell and Ian Osband and Ian Silber and Ian Sohl and Ibrahim Okuyucu and Ikai Lan and Ilya Kostrikov and Ilya Sutskever and Ingmar Kanitscheider and Ishaan Gulrajani and Jacob Coxon and Jacob Menick and Jakub Pachocki and James Aung and James Betker and James Crooks and James Lennon and Jamie Kiros and Jan Leike and Jane Park and Jason Kwon and Jason Phang and Jason Teplitz and Jason Wei and Jason Wolfe and Jay Chen and Jeff Harris and Jenia Varavva and Jessica Gan Lee and Jessica Shieh and Ji Lin and Jiahui Yu and Jiayi Weng and Jie Tang and Jieqi Yu and Joanne Jang and Joaquin Quinonero Candela and Joe Beutler and Joe Landers and Joel Parish and Johannes Heidecke and John Schulman and Jonathan Lachman and Jonathan McKay and Jonathan Uesato and Jonathan Ward and Jong Wook Kim and Joost Huizinga and Jordan Sitkin and Jos Kraaijeveld and Josh Gross and Josh Kaplan and Josh Snyder and Joshua Achiam and Joy Jiao and Joyce Lee and Juntang Zhuang and Justyn Harriman and Kai Fricke and Kai Hayashi and Karan Singhal and Katy Shi and Kavin Karthik and Kayla Wood and Kendra Rimbach and Kenny Hsu and Kenny Nguyen and Keren Gu-Lemberg and Kevin Button and Kevin Liu and Kiel Howe and Krithika Muthukumar and Kyle Luther and Lama Ahmad and Larry Kai and Lauren Itow and Lauren Workman and Leher Pathak and Leo Chen and Li Jing and Lia Guy and Liam Fedus and Liang Zhou and Lien Mamitsuka and Lilian Weng and Lindsay McCallum and Lindsey Held and Long Ouyang and Louis Feuvrier and Lu Zhang and Lukas Kondraciuk and Lukasz Kaiser and Luke Hewitt and Luke Metz and Lyric Doshi and Mada Aflak and Maddie Simens and Madelaine Boyd and Madeleine Thompson and Marat Dukhan and Mark Chen and Mark Gray and Mark Hudnall and Marvin Zhang and Marwan Aljubeh and Mateusz Litwin and Matthew Zeng and Max Johnson and Maya Shetty and Mayank Gupta and Meghan Shah and Mehmet Yatbaz and Meng Jia Yang and Mengchao Zhong and Mia Glaese and Mianna Chen and Michael Janner and Michael Lampe and Michael Petrov and Michael Wu and Michele Wang and Michelle Fradin and Michelle Pokrass and Miguel Castro and Miguel Oom Temudo de Castro and Mikhail Pavlov and Miles Brundage and Miles Wang and Minal Khan and Mira Murati and Mo Bavarian and Molly Lin and Murat Yesildal and Nacho Soto and Natalia Gimelshein and Natalie Cone and Natalie Staudacher and Natalie Summers and Natan LaFontaine and Neil Chowdhury and Nick Ryder and Nick Stathas and Nick Turley and Nik Tezak and Niko Felix and Nithanth Kudige and Nitish Keskar and Noah Deutsch and Noel Bundick and Nora Puckett and Ofir Nachum and Ola Okelola and Oleg Boiko and Oleg Murk and Oliver Jaffe and Olivia Watkins and Olivier Godement and Owen Campbell-Moore and Patrick Chao and Paul McMillan and Pavel Belov and Peng Su and Peter Bak and Peter Bakkum and Peter Deng and Peter Dolan and Peter Hoeschele and Peter Welinder and Phil Tillet and Philip Pronin and Philippe Tillet and Prafulla Dhariwal and Qiming Yuan and Rachel Dias and Rachel Lim and Rahul Arora and Rajan Troll and Randall Lin and Rapha Gontijo Lopes and Raul Puri and Reah Miyara and Reimar Leike and Renaud Gaubert and Reza Zamani and Ricky Wang and Rob Donnelly and Rob Honsby and Rocky Smith and Rohan Sahai and Rohit Ramchandani and Romain Huet and Rory Carmichael and Rowan Zellers and Roy Chen and Ruby Chen and Ruslan Nigmatullin and Ryan Cheu and Saachi Jain and Sam Altman and Sam Schoenholz and Sam Toizer and Samuel Miserendino and Sandhini Agarwal and Sara Culver and Scott Ethersmith and Scott Gray and Sean Grove and Sean Metzger and Shamez Hermani and Shantanu Jain and Shengjia Zhao and Sherwin Wu and Shino Jomoto and Shirong Wu and Shuaiqi and Xia and Sonia Phene and Spencer Papay and Srinivas Narayanan and Steve Coffey and Steve Lee and Stewart Hall and Suchir Balaji and Tal Broda and Tal Stramer and Tao Xu and Tarun Gogineni and Taya Christianson and Ted Sanders and Tejal Patwardhan and Thomas Cunninghman and Thomas Degry and Thomas Dimson and Thomas Raoux and Thomas Shadwell and Tianhao Zheng and Todd Underwood and Todor Markov and Toki Sherbakov and Tom Rubin and Tom Stasi and Tomer Kaftan and Tristan Heywood and Troy Peterson and Tyce Walters and Tyna Eloundou and Valerie Qi and Veit Moeller and Vinnie Monaco and Vishal Kuo and Vlad Fomenko and Wayne Chang and Weiyi Zheng and Wenda Zhou and Wesam Manassra and Will Sheu and Wojciech Zaremba and Yash Patil and Yilei Qian and Yongjik Kim and Youlong Cheng and Yu Zhang and Yuchen He and Yuchen Zhang and Yujia Jin and Yunxing Dai and Yury Malkov},
      year={2024},
      eprint={2410.21276},
      archivePrefix={arXiv},
      primaryClass={cs.CL},
      url={https://arxiv.org/abs/2410.21276}, 
}

@misc{ren2024grounding,
      title={Grounding DINO 1.5: Advance the "Edge" of Open-Set Object Detection}, 
      author={Tianhe Ren and Qing Jiang and Shilong Liu and Zhaoyang Zeng and Wenlong Liu and Han Gao and Hongjie Huang and Zhengyu Ma and Xiaoke Jiang and Yihao Chen and Yuda Xiong and Hao Zhang and Feng Li and Peijun Tang and Kent Yu and Lei Zhang},
      year={2024},
      eprint={2405.10300},
      archivePrefix={arXiv},
      primaryClass={cs.CV}
}

@misc{loshchilov2019decoupledweightdecayregularization,
      title={Decoupled Weight Decay Regularization}, 
      author={Ilya Loshchilov and Frank Hutter},
      year={2019},
      eprint={1711.05101},
      archivePrefix={arXiv},
      primaryClass={cs.LG},
      url={https://arxiv.org/abs/1711.05101}, 
}

@inproceedings{DeBortoli2021DiffusionSchrodingerBridge,
  author    = {Valentin De Bortoli and James Thornton and Jeremy Heng and Arnaud Doucet},
  title     = {Diffusion Schrödinger Bridge with Applications to Score-Based Generative Modelling},
  booktitle =NIPS,
  year      = {2021},
}

@inproceedings{chen2022likelihood,
  title={Likelihood Training of Schr{\"o}dinger Bridge using Forward-Backward SDEs Theory},
  author={Chen, Tianrong and Liu, Guan-Horng and Theodorou, Evangelos A},
  booktitle=ICLR,
  year={2022}
}

@article{Tong2024ImprovingCFM,
  author    = {Alexander Tong and Kilian Fatras and Nikolay Malkin and Guillaume Huguet and Yanlei Zhang and Jarrid Rector-Brooks and Guy Wolf and Yoshua Bengio},
  title     = {Improving and Generalizing Flow-Based Generative Models with Minibatch Optimal Transport},
  journal   = {Transactions on Machine Learning Research},
  year      = {2024},
  note      = {arXiv preprint arXiv:2302.00482},
  url       = {https://arxiv.org/abs/2302.00482}
}

@inproceedings{peebles2023scalable,
  title={Scalable diffusion models with transformers},
  author={Peebles, William and Xie, Saining},
  booktitle={Proceedings of the IEEE/CVF international conference on computer vision},
  pages={4195--4205},
  year={2023}
}

@inproceedings{saito2024soundctm,
    title={Sound{CTM}: Unifying Score-based and Consistency Models for Full-band Text-to-Sound Generation},
    author={Koichi Saito and Dongjun Kim and Takashi Shibuya and Chieh-Hsin Lai and Zhi Zhong and Yuhta Takida and Yuki Mitsufuji},
    booktitle=ICLR,
    year={2025},
    url={https://openreview.net/forum?id=KrK6zXbjfO}
}

@inproceedings{ronneberger2015u,
  title={U-net: Convolutional networks for biomedical image segmentation},
  author={Ronneberger, Olaf and Fischer, Philipp and Brox, Thomas},
  booktitle={International Conference on Medical image computing and computer-assisted intervention},
  pages={234--241},
  year={2015},
  organization={Springer}
}

@article{kong2020hifi,
  title={Hifi-gan: Generative adversarial networks for efficient and high fidelity speech synthesis},
  author={Kong, Jungil and Kim, Jaehyeon and Bae, Jaekyoung},
  journal={Advances in neural information processing systems},
  volume={33},
  pages={17022--17033},
  year={2020}
}

@article{loshchilov2017decoupled,
  title={Decoupled weight decay regularization},
  author={Loshchilov, Ilya and Hutter, Frank},
  journal={arXiv preprint arXiv:1711.05101},
  year={2017}
}

@article{lipman2022flow,
  title={Flow matching for generative modeling},
  author={Lipman, Yaron and Chen, Ricky TQ and Ben-Hamu, Heli and Nickel, Maximilian and Le, Matt},
  journal={arXiv preprint arXiv:2210.02747},
  year={2022}
}

@article{liu2022flow,
  title={Flow straight and fast: Learning to generate and transfer data with rectified flow},
  author={Liu, Xingchao and Gong, Chengyue and Liu, Qiang},
  journal={arXiv preprint arXiv:2209.03003},
  year={2022}
}

@article{song2020denoising,
  title={Denoising diffusion implicit models},
  author={Song, Jiaming and Meng, Chenlin and Ermon, Stefano},
  journal={arXiv preprint arXiv:2010.02502},
  year={2020}
}
}
\clearpage
\appendix

% Keep this: one entry in the main TOC
% \section*{Supplementary Material}\label{sec:supp}
\addcontentsline{toc}{section}{Supplementary Material}

% Start a partial TOC from here
\startcontents[appendix]

\setcounter{page}{1}
\maketitlesupplementary

% ===== Contents ONLY for appendix (sections A, B, ...) =====
\section*{Contents}
%   [appendix]   = which partial TOC
%   {}           = no special prefix
%   {1}          = start at level 1 (section)
%   [2]          = include down to depth 2 (section + subsection)
\printcontents[appendix]{}{1}[2]{}
\vspace{1.5em}

% We notice that in Grounding Dino demo page they include a period at the end of their text prompt. In our case, we only include one object. In order to test whether adding a period will result in different performance given the same threshold. We randomly select 50 frames from our dataset and report the result as follows.\weihantodo{manually check}

% \section{Demo Page}
% We provide a demo page under index.md with paths to representative examples and include folders containing all dataset pairs and editing pairs in the supplementary materials.

\section{Derivations for Schr\"{o}dinger Bridge Flow Matching}\label{sec:details_sb_math}
\label{app:sb}
\subsection{Preliminary: Probability flow}
For a stochastic differential equation (SDE)
\begin{align}
    \dd X_t=f_t(X_t)\dd t+g_t \dd W_t,
\end{align}
the equivalent \emph{Fokker-Planck equation} (FPE) is
\begin{align}
    \frac{\partial p_t}{\partial t} = -\nabla \cdot (f_t p_t) + \frac{1}{2}g_t^2 \Delta p_t,
\end{align}
which can be re-arranged into the following form of a \emph{continuity equation}:
\begin{align}\label{eq:pf}
    \frac{\partial p_t}{\partial t} = -\nabla \cdot p_t\underbrace{\left(f_t-\frac{1}{2}g_t^2\nabla \log p_t\right)}_{=:u_t},
\end{align}
where $u_t$ is the \emph{probability flow} (PF) of the SDE. The PF allows one to sample from the SDE \emph{deterministically} since it induces the same FPE as the original SDE. Also, notice that the probability flow can be uniquely determined once we know the SDE functions $(f_t,g_t)$ and the score function $\nabla\log p_t$.

\subsection{Solving Schr\"{o}dinger Bridge (SB)}
The solution to SB can be expressed by the following forward and backward stochastic differential equations (SDEs)~\cite{DeBortoli2021DiffusionSchrodingerBridge,chen2022likelihood}:
\begin{align}\label{eq:sb_sde}
    \dd X_t &= [f_t+\beta_t\nabla\log\Psi_t]\dd t+\sqrt{\beta_t}\dd W_t,\,X_0\sim p_0,\\
    \dd X_t &= [f_t+\beta_t\nabla\log\hat{\Psi}_t]\dd t+\sqrt{\beta_t}\dd \overline{W}_t,X_1\sim p_1,
\end{align}
where $f_t(X_t)\in \sR^d$ controls the drift, $\beta_t\in\sR$ is the diffusion, and $W_t,\overline{W}_t\in\sR^d$ are standard forward and backward Wiener processes, respectively. The additional drift terms $\nabla \log \Psi_t(X_t,y)$ and $\nabla\log \hat{\Psi}_t(X_t,y)$ are score functions satisfying the following coupled partial differential equations (PDE):
\begin{equation}
\begin{aligned}
    &\begin{cases}
        \frac{\partial\Psi}{\partial t}=-\nabla\Psi^\top f-\frac{1}{2}\beta\Delta\Psi\\
        \frac{\partial\hat{\Psi}}{\partial t}=-\nabla\cdot(\hat{\Psi} f)+\frac{1}{2}\beta\Delta\hat{\Psi}
    \end{cases},\Psi_0\hat{\Psi}_0=p_0,\Psi_1\hat{\Psi}_1=p_1.
\end{aligned}
\end{equation}
when $f_t\equiv 0$ and $\beta_t\equiv \sigma^2.$ 
%Therefore, as shown in \cite{Tong2024ImprovingCFM}, the conditional flow of SB has the following closed-form expression: 
%\paragraph{Learning Schr\"{o}dinger's bridge (SB) with paired data}
%\begin{align}\label{eq:sb_prob}
%    X_t|x_0,x_1\sim \mathcal{N}\left(tx_1+(1-t)x_0;\sigma^2t(1-t)\right).
%\end{align}
%\begin{align}\label{eq:sb_cond_flow}
%    u_t(X_t|X_0,X_1)=u_t^{\mathrm{CFM}}+\frac{1-2t}{2t(1-t)}(X_t-\mu_t),
%\end{align}
%where $X_t\sim \gN(\mu_t;\sigma^2t(1-t)I_d)$, which is pinched on $X_0$ and $X_1$ at the starting and end times, respectively. 
% Note that the second term in Eq.~\ref{eq:sb_cond_flow} is a correction term to the sampling path in the classical CFM to account for the stochasticity of $X_t$ due to $\sigma>0$, since we observe that a nonzero $\sigma$ leads to better generalization performance for Audiovisual Removal.

\subsection{Learning Schr\"{o}dinger Bridge (SB) with paired data}
% \liming{use $\beta_t$ instead of $g_t$ to be consistent with prior works}
A Schr\"{o}dinger bridge process $\{X_t\}_{t\in[0,1]}$ as defined in Eq.~\ref{eq:sb_sde} can be simulated once we have access to the score function $\nabla\log\Psi_t$ and the initial distribution $p_0$. In the presence of paired data $(x_0,x_1)$, this can be achieved via conditional score matching with the score of the conditional distribution $p_{t|01}$, similar to a diffusion model~\cite{sohl-dickstein2015deep,Song2019generative,Song2019sliced,Song2020improved,ho2020denoising,Song2021score}. To achieve this, we simulate from the following SDE:
\begin{align}\label{eq:sb_cond_sde}
    \dd X_t = \frac{x_1-X_t}{1-t}\dd t + \sigma \dd W_t,\,X_0=x_0,X_1=x_1,
\end{align}
which has the following closed-form expression for the marginal distributions:
\begin{align}\label{eq:sb_prob}
    X_t|x_0,x_1\sim \mathcal{N}\left(tx_1+(1-t)x_0;\sigma^2t(1-t)\right).
\end{align}
Note that the mean $\mu_t:=tx_1+(1-t)x_0$ of the process is an interpolation between the start and end points, and the variances of the process at start $(t=0)$ and end $(t=1)$ times are both 0, consistent with the fixed-point constraints imposed.

\subsection{Learning SB via flow matching}
Instead of performing score matching, we can derive the probability flow of Eq.~\ref{eq:sb_sde} by applying Eq.~\ref{eq:pf} with $f_t(x)=\frac{x_1-x}{1-t}$ and $g_t=\sigma$:
\begin{align}
    u_t(x|x_0,x_1)&=\frac{x_1-x}{1-t}-\frac{\sigma^2}{2}\nabla \log p_{t|01}(x|x_0,x_1)\nonumber\\
    &=\frac{x_1-x}{1-t}-\frac{\mu_t-x}{2t(1-t)}\nonumber\\
    &=(t-1/2)\frac{\mu_t-x}{t(1-t)}+\frac{x_1-\mu_t}{1-t}\nonumber\\
    &=\frac{1-2t}{2t(1-t)}(x-\mu_t)+(x_1-x_0).
\end{align}
Note that the second term is the same as the conditional flow for the vanilla flow matching~\cite{lipman2023flow}, as it is essentially a special case of the SB flow by setting the noise schedule parameter to be $\sigma\equiv0$.

\section{Dataset Details}\label{sec:dataset_details}

\subsection{Prompts}
\paragraph{Sounding Object Detection}
We first use Qwen-VL~\cite{Qwen-VL} with the prompt:
\textit{``What are the sounding sources appearing in the video?''}
We then use GPT-4o-mini~\cite{openai2024gpt4ocard} to distill Qwen’s response with the prompt:
\textit{``List all entities mentioned in this caption that can produce sound on their own. Return only the entity names as a comma-separated list.''}

\paragraph{Video-to-Audio Generation}
We use MMAudio~\cite{cheng2025taming} to generate an audio track for each object, using the visual as a visual cue as well as the object name as a text prompt. We set the CFG strength to 6 and run inference for 50 steps.

\paragraph{Audio Quality Verification}
We use Qwen-Audio~\cite{Qwen-Audio} to verify the quality of the generated audio. Given all predicted object names from our sounding object detection stage and their corresponding audio tracks, we ask Qwen-Audio to choose the most likely sounding object and to estimate how many sound sources are present in the audio. We then use GPT-4o-mini~\cite{openai2024gpt4ocard} to distill Qwen-Audio’s answer with the prompt:
\textit{``You are given an answer from an audio model:\textbackslash n\{ans\}\textbackslash n Return ONLY the object name, nothing else. Do not include quotes, punctuation, or explanatory text.''}
We retain only audio tracks that contain exactly one sound source, and this source satisfies the condition specified in our video-to-audio generation stage.

\subsection{Object Categorization}\label{sec:object_category}
In this section, we show how we categorize each object via simple, case-insensitive keyword matching on its name (we lowercase the string and check for substrings). Note that in our dataset annotation, we use its original object name without categorization.
The first matching group in the following fixed order is assigned: \emph{People/Human} (e.g., man, woman, person, boy, girl, child, baby, human, voice, speech, mother, father, crowd, people, toddler, singer); \emph{Vehicles} (car, truck, motorcycle, airplane/plane, train, boat/ship, helicopter, vehicle, tractor, bus, bicycle, scooter); \emph{Electronics/Devices} (keyboard, computer, tv, phone, microphone, speaker, laptop, camera, screen/monitor, tablet, radio, device); \emph{Engines/Machinery} (engine, excavator, machinery, bulldozer, motor, loader, drill, grinder, machine, mechanical); \emph{Musical Instruments} (guitar, piano, violin, drum, saxophone, trumpet, instrument, accordion, flute, cello, banjo, horn, sitar, marimba); \emph{Animals} (dog, cat, bird, cow, horse, lion, duck, sheep, pig, animal, tiger, elephant, bear, fish, owl); \emph{Household Items} (clock, faucet, refrigerator, door, sink, oven, microwave, washing, vacuum, fan, alarm, stove, toilet, chair, table); \emph{Tools/Implements} (knife, saw, hammer, drill, wrench, scissors, pen, pencil, tool, screwdriver, chisel, spoon, fork); and \emph{Nature/Environment} (rain, wind, fire, water, ocean, river, snow, storm, thunder, wave, leaf, tree, forest). If multiple categories match, we use the first match in this order.

\subsection{Dataset Scalability}\label{sec:dataset_scability}
\begin{figure}
    \centering
    \includegraphics[width=1\linewidth]{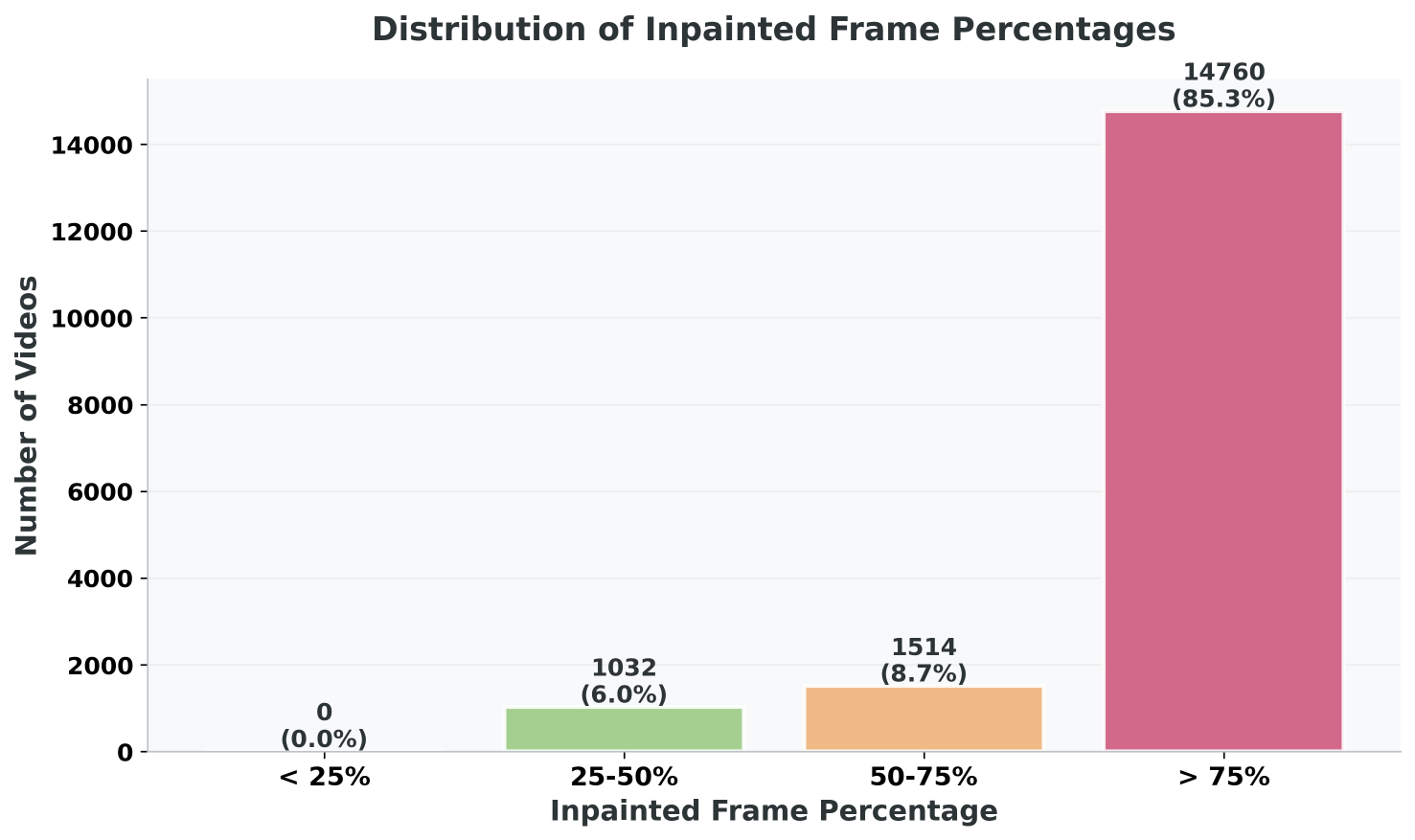}
    \caption{Number of inpainted frames out of all frames in each 8-second, 24-fps clip.}
    \label{fig:inpainted_details}
\end{figure}

The most time-consuming step is Grounding DINO~\cite{ren2024grounding} detection to obtain bounding boxes, followed by SAM2~\cite{ravi2024sam2segmentimages} to compute segmentation masks and remove the target object on individual frames. To accelerate processing, we run detection every \(k\) frames (e.g., \(k{=}24\)) and temporally propagate the boxes and masks in between. Because our target objects are often relatively small, we only keep samples where most frames containing the target object are inpainted. We show the distribution of the number of inpainted frames in Figure~\ref{fig:inpainted_details}, and report the retention rate and average processing time per sample on an RTX A6000 in Table~\ref{tab:pipeline_yield}. We report the retention rate as a percentage:
\[
\text{Retention Rate (\%)} = 100 \times \frac{\text{\# outputs}}{\text{\# inputs}}.
\]

\begin{table}[ht]
  \centering
  \small
  \setlength{\tabcolsep}{2pt} 
  \renewcommand{\arraystretch}{0.9} 
  \begin{tabular}{@{}lccc@{}}   
    \toprule
    \textbf{Stage} & \textbf{Retention Rate (\%)} & \textbf{Avg. Time Per Sample (s)} \\
    \midrule
    VLM &    92    &    $\approx$ 15 \\
    V2A synthesis  &    44    &   $\approx$ 25     \\
    Video Inpainting    &   49    &   $\approx$ 685     \\
    \bottomrule
  \end{tabular}
  \caption{Per-stage retention rate and runtime in our data pipeline.}
  \label{tab:pipeline_yield}
\end{table}

% We set strict bounding box threshold which makes the inpainting roughly low.

\section{Implementation Details}

\paragraph{Model Configuration}
For audio VAE, audio is sampled at 44.1 kHz. For video VAE, each frame is resized to 128×128. We configure the backbone main DiT with hidden dimension of 2048, comprising 16 layers and 16 attention heads. To effectively disentangle audio-visual mixtures, we use two small DiTs to serve as audio head and video head. Each small DiT is composed of 4 layers and 4 attention heads. For the sampling method, we use the Euler method to solve the Ordinary Differential Equation (ODE) with 30 sampling steps for both audio and video latent.
\paragraph{Training Configuration}
We train on four H100 GPUs for 12 hours and select the checkpoint with the lowest validation loss, which is the checkpoint at 42 epochs. We use the AdamW optimizer\cite{loshchilov2019decoupledweightdecayregularization} with $\beta_1 = 0.9$ and $\beta_2 = 0.999$. The initial learning rate is $10^{-5}$ with 5000 warmup steps to reach a learning rate of $10^{-4}$. The weight decay is $10^{-4}$. We also use a weighted loss. In particular, we define the loss as
\[
\text{loss} = \text{audio\_loss} + 3 \cdot \text{video\_loss}.
\]

\paragraph{Implementation Details for AUDIT}\label{sec:audit}
We re-implemented AUDIT \cite{wang2023audit} to serve as a baseline for the audio removal task.  The original model operates at a 16 kHz sample rate, utilizing a U-Net backbone \cite{ronneberger2015u} and vocoder-based \cite{kong2020hifi} audio generation. To ensure a fair and high-fidelity comparison, our re-implementation adopted a 44.1 kHz sample rate, leveraged the scalable DiT backbone \cite{peebles2023scalable}, and employed a waveform-based VAE \cite{saito2024soundctm}. Furthermore, for the sampling process, instead of using DDIM \cite{song2020denoising}, we utilize conditional flow matching (CFM) \cite{lipman2022flow, liu2022flow}, specifically employing the Euler method with 30 inference steps. Training was performed for 300 epochs on one NVIDIA Blackwell A6000 GPU. The training utilized the AdamW optimizer \cite{loshchilov2017decoupled} with a batch size of 64, a two-stage learning rate schedule ($10^{-5}$ initial rate with 5000 warmup steps, then $10^{-4}$), and optimizer hyperparameters $\beta_1=0.9$, $\beta_2=0.999$, $\epsilon=10^{-8}$, and a weight decay of $10^{-4}$.

\begin{table*}[t]
  \centering
  \scriptsize
  \setlength{\tabcolsep}{3pt}
  \renewcommand{\arraystretch}{0.9}
  \begin{tabular}{llccccc}
    \toprule
    & &
    \multicolumn{2}{c}{\textbf{Prompt Following$\uparrow$}} 
    & \multicolumn{2}{c}{\textbf{Output Fidelity$\uparrow$}} 
    & \multicolumn{1}{c}{\textbf{ Synchronization$\uparrow$}} \\
    \cmidrule(lr){3-4}\cmidrule(lr){5-6}\cmidrule(lr){7-7}
    \textbf{Video Model} 
      & \textbf{Audio Model}
      & \textbf{Audio} 
      & \textbf{Video}
      & \textbf{Audio} 
      & \textbf{Video}
      & \textbf{Audiovisual} \\
    \midrule
    Oracle$^{*}$ & Oracle$^{*}$ 
      & 3.22$\pm$0.20 & 3.35$\pm$0.18 & 3.44$\pm$0.16 & 3.07$\pm$0.16 &  3.26$\pm$0.16\\
      \midrule
    VACE~\cite{jiang2025vaceallinonevideocreation} & ZEUS~\cite{manor2024zeroshotunsupervisedtextbasedaudio}  
      & 2.30$\pm$0.35 & 2.22$\pm$0.39 & 2.86$\pm$0.35  & \textbf{3.00$\pm$0.39} & 2.80$\pm$0.33 \\
    VideoPainter\cite{bian2025videopainteranylengthvideoinpainting} & ZEUS  
      & 2.17$\pm$0.31 & 2.61$\pm$0.35 & 2.53$\pm$0.29 & 1.80$\pm$0.24  & 2.22$\pm$0.25 \\
    % VideoPainter~\cite{bian2025videopainteranylengthvideoinpainting} & AUDIT &  & &  &  &  \\
    MCFM & MCFM 
      & \textbf{3.18$\pm$0.39} & \textbf{3.24$\pm$0.43} & 1.82$\pm$0.33 & 1.16$\pm$0.14  & 1.47$\pm$0.18 \\

    \midrule
    SAVE & SAVE 
      & 2.04$\pm$0.29 & 2.20$\pm$0.29 & \textbf{2.92$\pm$0.24} & 2.88$\pm$0.24  & \textbf{2.82$\pm$0.24} \\
    \bottomrule
    \multicolumn{7}{c}{\scriptsize $^{*}$ Synthetic audiovisual pairs from our preprocessing pipeline, constructed using pretrained models,} \\
    \multicolumn{7}{c}{\scriptsize are treated as oracle annotations that approximate an upper bound.} \\
  \end{tabular}
  \caption{\textbf{Subjective evaluation} (mean $\pm$ 95\% confidence interval over participants).}
  \label{tab:subjective_eval}
\end{table*}

\begin{table*}[t]
  \centering
  \scriptsize
  \setlength{\tabcolsep}{3pt}
  \renewcommand{\arraystretch}{0.9}
  \begin{tabular}{lcccccccc}
    \toprule
    & \multicolumn{4}{c}{\textbf{Video}} & \multicolumn{2}{c}{\textbf{Audio}} & \multicolumn{2}{c}{\textbf{Audiovisual Synchronization}} \\
    \cmidrule(lr){2-5}\cmidrule(lr){6-7}\cmidrule(lr){8-9}
    \textbf{Condition}
      & \textbf{PSNR$\uparrow$} & \textbf{SSIM$\uparrow$} & \textbf{LPIPS$\downarrow$} & \textbf{FVID$\downarrow$}
      & \textbf{FAD$\downarrow$} & \textbf{LSD$\downarrow$}
      & \textbf{DeSync$\downarrow$} & \textbf{ImageBind$\uparrow$} \\
    \midrule
    Semantic Mask          & \textbf{20.82$\pm$5.27} & \textbf{0.69$\pm$0.16}  & 0.02$\pm$0.01 & \textbf{13.95} & 0.71 & 1.46 & 0.84 & 0.21  \\
    Semantic Mask \& CLAP  & 20.43 $\pm$ 5.25 & 0.68$\pm$0.17 & 0.02$\pm$0.01 & 14.27 & \textbf{0.67} &  \textbf{1.44} & \textbf{0.83} & 0.21  \\
    \bottomrule
  \end{tabular}
  \caption{\textbf{Effect of CLAP conditioning.} Comparison of video quality, audio quality, and audiovisual synchronization under different conditioning schemes.}
  \label{tab:clap_condition_ablation}
\end{table*}

\section{Subjective Evaluation}\label{sec:subjective_test}
We recruit 15 participants to evaluate both the quality of our dataset construction and the editing performance of SAVE compared with our baselines. We ask them to rate each question on a 5-point Likert scale, where 1 means “not at all” / “very bad” and 5 means “very likely” / “very good.” We split the study into two versions: Version A with 7 participants and Version B with 8 participants. We present 10 constructed dataset pairs in Version A, 12 in Version B, and 4 editing sample pairs for each version.

From our objective evaluation, VACE~\cite{jiang2025vaceallinonevideocreation} achieves the second-highest video quality after removal, while ZEUS~\cite{manor2024zeroshotunsupervisedtextbasedaudio} achieves the second-highest audio quality after removal. In addition, MCFM (Multimodal Conditional Flow Matching) obtains the second-lowest DeSyncScore, an objective metric where lower values indicate less temporal misalignment between audio and video, followed by VideoPainter~\cite{bian2025videopainteranylengthvideoinpainting} and ZEUS. Therefore, we focus on these three cross-modal combinations as baselines and report their subjective evaluation results in Table~\ref{tab:subjective_eval}.

We report the mean and 95\% confidence interval in Table~\ref{tab:subjective_eval}. We can see that although MCFM has the highest removal performance, its output fidelity for both modalities is very low because most of the outputs are random noise. The combination of VACE and ZEUS achieves the highest output video quality, while our proposed model achieves the highest output audio quality. In addition, our proposed model achieves the best audiovisual synchronization.

\subsection{Dataset Quality}\label{sec:dataset_quality}
We evaluate our constructed dataset by providing the source video and asking participants to evaluate our constructed pairs from the perspectives of audio prompt following, video prompt following, audio output fidelity, video output fidelity, and audiovisual synchronization.  We ask the following questions and treat this as the orcale for our model and report results in Table~\ref{tab:subjective_eval}. 
\begin{itemize}
    \item \textbf{Audio Prompt Following:} Given the input and output, how likely is it that the target object has been successfully removed from the audio?
    \item \textbf{Video Prompt Following:} Given the input and output, how likely is it that the target object has been successfully removed from the video?
    \item \textbf{Audio Output Fidelity:} Compare with the source sounding video, how would you rate the overall quality of this audio?
    \item \textbf{Video Output Fidelity:} Compare with the source sounding video, how would you rate the overall quality of this video?
    \item \textbf{Audiovisual Synchronization:} Compare with the source sounding video, how would you rate the synchronization between the generated audio and visual content?
\end{itemize}

\subsection{SAVE Performance}\label{sec:save_quality}
We evaluate our audiovisual editor by providing the source video, the outputs from our proposed audiovisual editor SAVE, as well as several baseline models, and asking participants to evaluate the edited pairs from the perspectives of audio prompt following, video prompt following, audio output fidelity, video output fidelity, and audiovisual synchronization. We ask the following questions and present our result in Table~\ref{tab:subjective_eval}.
\begin{itemize}
    \item \textbf{Audio Prompt Following:} Given the input and output, how likely is it that the target object has been successfully removed from the audio?
    \item \textbf{Video Prompt Following:} Given the input and output, how likely is it that the target object has been successfully removed from the video?
    \item \textbf{Audio Output Fidelity:} Compare with the source sounding video, how would you rate the overall quality of this audio?
    \item \textbf{Video Output Fidelity:} Compare with the source sounding video, how would you rate the overall quality of this video?
    \item \textbf{Audiovisual Synchronization:} Compare with the source sounding video, how would you rate the synchronization between the generated audio and visual content?
\end{itemize}

% \begin{table*}[t]
%   \centering
%   \scriptsize
%   \setlength{\tabcolsep}{3pt}
%   \renewcommand{\arraystretch}{0.9}
%   \begin{tabular}{lcccccccc}
%     \toprule
%     & \multicolumn{4}{c}{\textbf{Video Eval}} & \multicolumn{2}{c}{\textbf{Audio Eval}} & \multicolumn{2}{c}{\textbf{AV Sync}} \\
%     \cmidrule(lr){2-5}\cmidrule(lr){6-7}\cmidrule(lr){8-9}
%     \textbf{Condition}
%       & \textbf{PSNR$\uparrow$} & \textbf{SSIM$\uparrow$} & \textbf{LPIPS$\downarrow$} & \textbf{FVID$\downarrow$}
%       & \textbf{FAD$\downarrow$} & \textbf{LSD$\downarrow$}
%       & \textbf{DeSync$\downarrow$} & \textbf{ImageBind$\uparrow$} \\
%     \midrule
%     Semantic Mask          & \textbf{20.82$\pm$5.27} & \textbf{0.69$\pm$0.16}  & 0.02$\pm$0.01 & \textbf{13.95} & 0.71 & 1.46 & 0.84 & 0.21  \\
%     Semantic Mask \& CLAP  & 20.43 $\pm$ 5.25 & 0.68$\pm$0.17 & 0.02$\pm$0.01 & 14.27 & \textbf{0.67} &  \textbf{1.44} & \textbf{0.83} & 0.21  \\
%     \bottomrule
%   \end{tabular}
%   \caption{\textbf{Effect of CLAP conditioning.} Comparison of video quality, audio quality, and audiovisual synchronization under different conditioning schemes.}
%   \label{tab:clap_condition_ablation}
% \end{table*}

% WARNING: do not forget to delete the supplementary pages from your submission 
% \input{sec/X_suppl}

\end{document}